\pdfoutput=1

\documentclass[11pt]{article}

\usepackage[final]{acl}

\usepackage{times}
\usepackage{latexsym}

\usepackage[T1]{fontenc}

\usepackage[utf8]{inputenc}

\usepackage{microtype}

\usepackage{inconsolata}

\usepackage{amssymb}%
\usepackage{pifont}%
\usepackage{graphicx}
\usepackage{hyperref}
\usepackage{url}
\usepackage{microtype}
\usepackage{times}
\usepackage{latexsym}
\usepackage{soul}
\usepackage{amsmath}
\usepackage{array, booktabs}
\usepackage{bm}
\usepackage{cleveref}
\usepackage{graphicx}
\usepackage{tabularx}
\usepackage{soul}
\usepackage{xcolor}
\usepackage{todonotes}
\usepackage{multirow}
\usepackage{xpatch}
\usepackage{blindtext}
\usepackage{fdsymbol}
\usepackage{microtype}
\usepackage{hyperref}
\usepackage[export]{adjustbox}
\usepackage{textcomp}
\usepackage{xparse}
\usepackage{enumitem}
\usepackage{makecell}
\usepackage{subcaption}
\usepackage{colortbl}
\usepackage{tcolorbox}
\usepackage{xparse}
\usepackage{stfloats}
\usepackage{highlight}

\usepackage[linesnumbered,vlined,ruled]{algorithm2e}

\newcolumntype{P}[1]{>{\centering\arraybackslash}p{#1}}
\newcommand{\header}[1]{\text{#1}}
\newcommand{\modelname}[1]{\texttt{#1}}

\newcommand{\benchmark}[1]{\texttt{CRMArena}}
\newcommand{\datagen}[1]{\textsc{SyntheticCRM}}

\newcommand{\cmark}{\ding{51}}%
\newcommand{\xmark}{\ding{55}}%
\definecolor{c2}{RGB}{218,0,0}

\definecolor{lightblue}{RGB}{212, 235, 255}
\definecolor{lightorange}{RGB}{255, 204, 168}
\definecolor{lightyellow}{RGB}{255, 255, 168}
\definecolor{lightred}{RGB}{255, 168, 168}
\definecolor{darkred}{RGB}{234, 107, 102}
\definecolor{darkerblue}{RGB}{103, 136, 184}
\definecolor{lightgreen}{RGB}{144, 238, 144}

\definecolor{gold}{rgb}{0.83, 0.69, 0.22}
\sethlcolor{lightblue}
\newcolumntype{Y}{>{\centering\arraybackslash}X}

\NewDocumentCommand{\steeve}
{ mO{} }{\textcolor{gold}{\textsuperscript{\textit{Steeve}}\textsf{\textbf{\small[#1]}}}}
\NewDocumentCommand{\jason}
{ mO{} }{\textcolor{red}{\textsuperscript{\textit{Jason}}\textsf{\textbf{\small[#1]}}}}

\NewDocumentCommand{\alex}
{ mO{} }{\textcolor{blue}{\textsuperscript{\textit{Alex}}\textsf{\textbf{\small[#1]}}}}
\NewDocumentCommand{\sid}
{ mO{} }{\textcolor{green}{\textsuperscript{\textit{Sid}}\textsf{\textbf{\small[#1]}}}}
\NewDocumentCommand{\philippe}
{ mO{} }{\textcolor{magenta}{\textsuperscript{\textit{Philippe}}\textsf{\textbf{\small[#1]}}}}
\NewDocumentCommand{\yixin}
{ mO{} }{\textcolor{violet}{\textsuperscript{\textit{Yixin}}\textsf{\textbf{\small[#1]}}}}
\NewDocumentCommand{\caiming}
{ mO{} }{\textcolor{orange}{\textsuperscript{\textit{Caiming}}\textsf{\textbf{\small[#1]}}}}
\NewDocumentCommand{\ap}
{ mO{} }{\textcolor{brown}{\textsuperscript{\textit{Akshara}}\textsf{\textbf{\small[#1]}}}}

\definecolor{gold}{rgb}{0.83, 0.69, 0.22}

\crefformat{section}{\S#2#1#3} %
\crefformat{subsection}{\S#2#1#3}
\crefformat{subsubsection}{\S#2#1#3}

\title{\includegraphics[width=1.5cm]{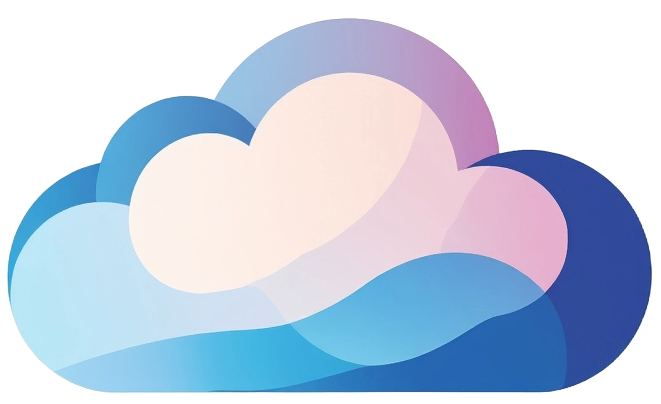}~ \benchmark~: Understanding the Capacity of LLM Agents to\\ Perform Professional CRM Tasks in Realistic Environments}

\author{Kung-Hsiang Huang ~~Akshara Prabhakar ~~Sidharth Dhawan ~~Yixin Mao\\ {\bfseries Huan Wang ~~Silvio Savarese ~~Caiming Xiong ~~Philippe Laban ~~Chien-Sheng Wu} \\
Salesforce AI Research\\
\texttt{\{kh.huang, akshara.prabhakar, sidharth, y.mao,} \\
\texttt{huan.wang, ssavarese, cxiong, wu.jason\}@salesforce.com} \\
}

\begin{document}
\maketitle

\begin{abstract}
Customer Relationship Management (CRM) systems are vital for modern enterprises, providing a foundation for managing customer interactions and data. 
Integrating AI agents into CRM systems can automate routine processes and enhance personalized service. 
However, deploying and evaluating these agents is challenging due to the lack of realistic benchmarks that reflect the complexity of real-world CRM tasks. 
To address this issue, we introduce \textbf{\benchmark~}, a novel benchmark designed to evaluate AI agents on realistic tasks grounded on professional work environments. 
We worked with CRM experts to design nine customer service tasks distributed across three personas: service agent, analyst, and manager. We synthesize a large-scale simulated organization, populating 16 commonly-used industrial objects (e.g., account, order, knowledge article, case) with high interconnectivity, and uploading it into a real Salesforce CRM organization. UI and API access to the CRM is provided to systems that attempt to complete the tasks in \benchmark~.
Experimental results reveal that state-of-the-art LLM agents succeed in less than 58\% of the tasks with ReAct prompting, and less than 65\% even when provided manually-crafted function-calling tools.
Our findings highlight the need for enhanced agent capabilities in function-calling and rule-following to be deployed in real-world work environment.
\benchmark~ is an open challenge to the community: systems that can reliably complete tasks showcase direct business value in a popular work environment.\footnote{Our code and benchmark have been released at \url{https://github.com/SalesforceAIResearch/CRMArena}.}\looseness=-1
\end{abstract}

\section{Introduction}

Customer Relationship Management (CRM) systems are pivotal in modern enterprises, serving as the backbone for managing interactions with current and potential customers~\cite{winer2001framework,payne2005strategic}. The integration of intelligent agents based on large language models (LLMs) into CRM systems promises to automate routine tasks, enhance operational efficiency, and revolutionize customer experiences. However, evaluating LLM agents in real-world professional environments remains a challenge, due to the absence of robust benchmarks that faithfully capture the complexity of tasks encountered in real-world CRM environments, largely due to data privacy concerns within enterprises.

Prior benchmarks on evaluating LLM agents on work-related tasks, such as WorkArena \cite{drouin2024workarena}, Workbench \cite{styles2024workbench}, and Tau \cite{yao2024tau} tend to focus on basic functionality, and fall short in two key areas. First, the complexity of the objects (e.g., tables in databases) and dependencies (e.g., foreign keys) between these objects is often overly simple, lacking the complexity of real-world scenarios. Second, the tasks included in the benchmarks, such as navigating web pages and filtering lists, are typically too straightforward and do not represent real-world work tasks. \looseness=-1

\begin{figure*}[t]
    \centering
    \includegraphics[width=0.85\linewidth]{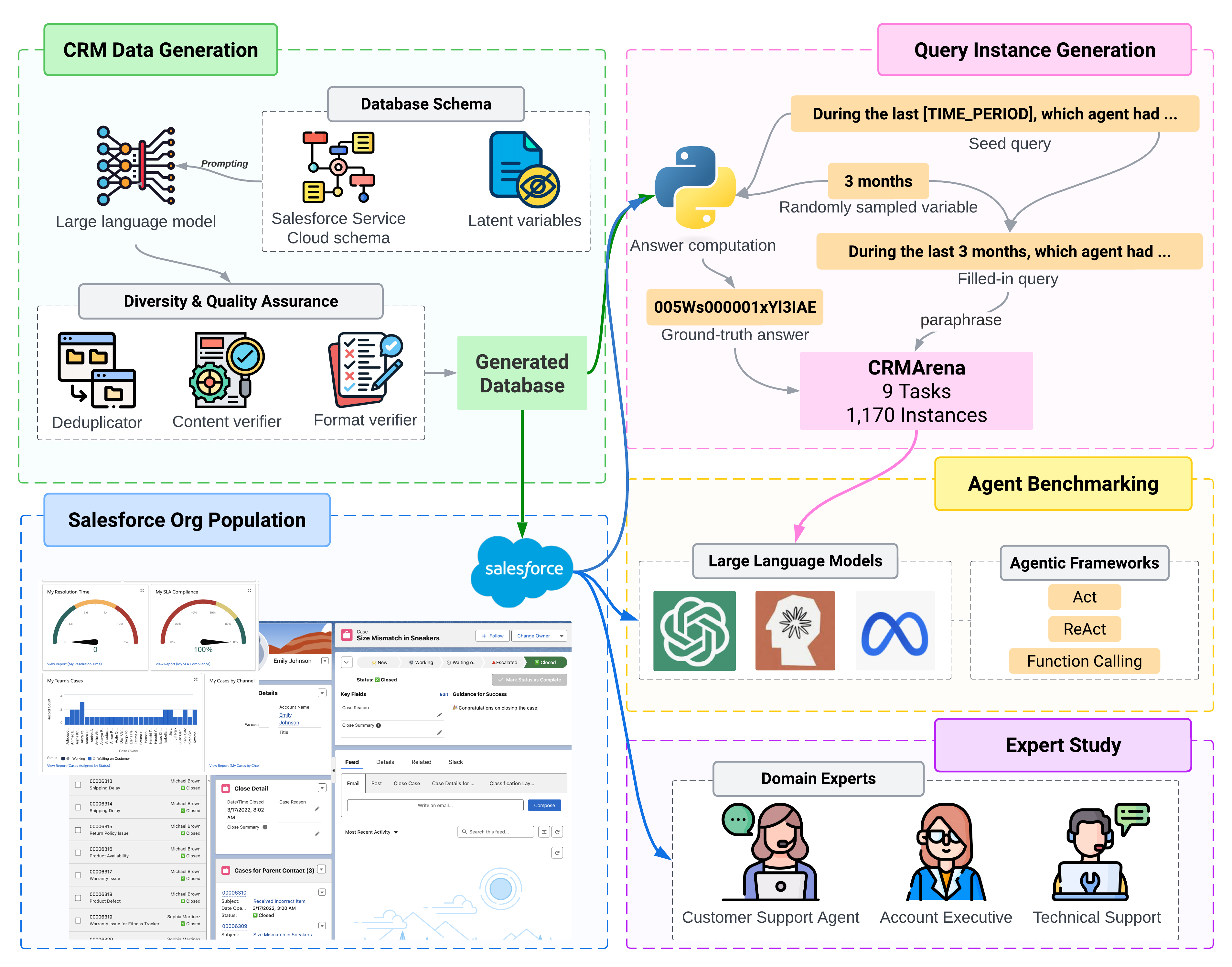}
    \vspace{-1mm}
    \caption{An overview of the contribution of this work. We begin by generating realistic CRM data based on the Salesforce Service Cloud schema, ensuring both quality and diversity through rigorous verification processes. This verified data is then stored locally and uploaded to a Salesforce organization (Org). An expert study, conducted with domain experts, validated the data's realism. Using this Org as a sandbox environment, we create query instances and benchmark various LLMs across different agentic frameworks.}
    \vspace{-3mm}
    \label{fig:project_overview}
\end{figure*}

To address these limitations, we introduce \textbf{\benchmark~}, a comprehensive benchmark tailored to evaluate LLM agents on performing realistic CRM tasks in real-world work environments. \benchmark~ features a realistic sandbox environment modeled after Salesforce’s schema, developed using an extensible data generation pipeline powered by LLMs (top left of \Cref{fig:project_overview}). Specifically, the pipeline tackles two key challenges: (1) \textit{Object connectivity}: reflecting the complex relationships between data objects (e.g., \textsc{Account} associated with \textsc{Case} and \textsc{Order}) by mirroring Salesforce's Service Cloud schema\footnote{\url{https://architect.salesforce.com/diagrams/data-models/service-cloud/service-cloud-overview}}. (2) Introducing \textit{latent variables} to better simulate realistic data dynamics, such as influencing case-filing behavior and modeling deviations from company guidelines.%

Moreover, \benchmark~ defines tasks based on actual customer service personas. By consulting CRM experts experienced with Salesforce, we identified nine tasks representative of CRM use cases (\Cref{sec:tasks}). These tasks span three personas: Service Manager, Service Agent, and Service Analyst. For example, Service Managers focus on agent performance and strategic resource allocation. \Cref{tab:dataset_comparison} compares \benchmark~ with previous datasets. \looseness=-1

\benchmark~ seamlessly integrates with Salesforce,\footnote{\url{https://www.salesforce.com/crm/}} enabling interaction via both the user interface and API access (see bottom of \Cref{fig:project_overview}). This integration facilitated an expert study with CRM professionals to assess the quality of our synthesized organization (\Cref{subsec:expert_study}). Study findings revealed that \textbf{90\% of domain experts found the test environment to be \textit{Realistic} or better}, underscoring the benchmark's fidelity to real-world CRM scenarios. %
Upon verifying the realism of \benchmark~, we then assess various agentic systems through API access. We develop two sets of tools \textit{general-purpose} vs. \textit{task-specific} tools, combine them with three agentic frameworks and various LLMs. Findings indicate that \textbf{all LLM agents struggle to reliably complete tasks when using general-purpose tools}, with top performing systems completing less than 40\% of the tasks. Incorporating manually designed tools can enhance performance, with top LLM agents solving up to 55\% of the tasks. However, we discover that \textbf{weaker LLMs often do not benefit from manually-crafted tools due to their weaker function calling capabilities}. %

In summary, our main contributions are: 
\begin{itemize}[leftmargin=*]\itemsep0em 
    \item Introducing \benchmark~, a realistic CRM agent benchmark with tasks validated by domain experts to evaluate LLM agents in real-world business scenarios.\looseness=-1
    \item Developing a data generation strategy anchored in a real-world CRM schema, incorporating latent variables, deduplication, and rigorous data validation to ensure diversity and quality.
    \item Demonstrating through experiments that even state-of-the-art LLM agents do not reliably complete \benchmark~ tasks, emphasizing the benchmark's value and challenges.
    
\end{itemize}

\section{\benchmark~}

Motivated by tasks commonly addressed by CRM personas: service manager, service agent, and service analyst, \benchmark~ comprises nine tasks that reflect real-world CRM scenarios. Verified by domain experts as common occurrences in CRM, an overview of these tasks is presented in \Cref{fig:task_overview}. Below, we provide detailed illustrations of each task. \looseness=-1

\subsection{Tasks}
\label{sec:tasks}

The tasks in \benchmark~ are designed to accurately reflect the variety of challenges encountered in real-world CRM environments. They span the responsibilities of three key personas: the Service Manager, who focuses on strategic resource allocation; the Service Agent, who addresses customer inquiries; and the Service Analyst, who analyzes data trends and performance metrics to improve service operations. \looseness=-1

\begin{table*}[t]
    \small
    \centering
    \begin{adjustbox}{max width=0.98\textwidth}
    {
    \begin{tabular}{lcccccc}
        \toprule
        
        \textbf{Datasets} & \textbf{\# Objects} & \textbf{\# Dependencies/ Object} & \textbf{Real-world Environment} & \textbf{Realistic Work Tasks} & \textbf{Expert Validation}\\
        \midrule
        
        WorkBench \cite{styles2024workbench} & 5 & 0 & {\color{darkred} \xmark } & {\color{darkred} \xmark } & {\color{darkred} \xmark }\\
        Tau-Bench \cite{yao2024tau} & 3 & 0.67 & {\color{darkred} \xmark } & {\color{darkred} \xmark } & {\color{darkred} \xmark }\\
        WorkArena \cite{drouin2024workarena}  & 7 & 0.86 & {\color{lightgreen} \cmark} & {\color{darkred} \xmark } & {\color{darkred} \xmark }\\
        \midrule

        \benchmark~ (Ours)  & \textbf{16} & \textbf{1.31} & {\color{lightgreen} \cmark} & {\color{lightgreen} \cmark} & {\color{lightgreen} \cmark}\\
        
        \bottomrule
    \end{tabular}
    }
    \end{adjustbox}
    \vspace{-1mm}
    
    \caption{\textbf{A comparison between our benchmark with prior datasets.} \benchmark~ is the most complex benchmark given the highest number of objects and object dependencies involved. Furthermore, \benchmark~ is the only expert-validated benchmark that encompasses both a real-world environment and realistic work tasks.} 
    \label{tab:dataset_comparison}
    \vspace{-3mm}
\end{table*}

\paragraph{New Case Routing (NCR)}
The goal of this task is to assign the best human agent to an incoming case, aiming to optimize various performance metrics. The input consists of a case subject and description%
, and the output is the ID of the recommended human agent. This task assesses LLM agent's %
ability to match cases to the most suitable human agent based on case histories and the skills and availability of these agents.%
\vspace{-5pt}
\paragraph{Handle Time Understanding (HTU)}
This task involves identifying the agent with the shortest/longest average handle time. Given the case history data, the objective is to determine the human agent who handled the previous cases the fastest/slowest. %
\vspace{-5pt}
\paragraph{Transfer Count Understanding (TCU)}
In this task, the LLM agent must find out which human agent %
transferred cases to others the least/most given a period of case history. Both HTU and TCU evaluate LLM agent's capacity to analyze performance based on predefined metrics accurately.

\begin{figure*}[t]
    \centering
    \includegraphics[width=0.98\linewidth]{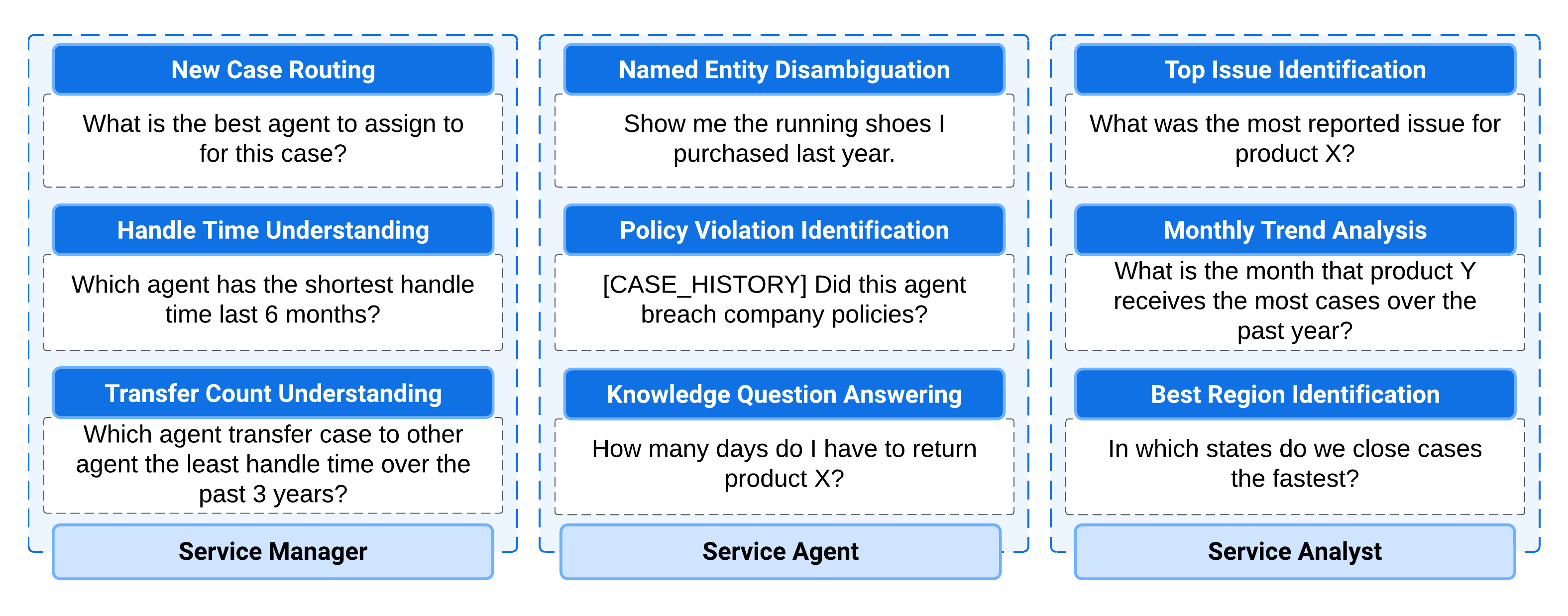}
    \vspace{-4mm}
    \caption{An overview of the nine distinct tasks introduced in \benchmark~.} %
    \vspace{-5mm}
    \label{fig:task_overview}
\end{figure*}

\vspace{-5pt}
\paragraph{Name Entity Disambiguation (NED)}
The LLM agent must disambiguate named entities related to customer transactions. Here, we focus on disambiguating product names. %
Given the query shown in \Cref{fig:task_overview}, the agent needs to identify the specific order corresponding to running shoes bought by the mentioned customer within the given time frame. This tests the understanding of product names and customer order histories.
\vspace{-5pt}
\paragraph{Policy Violation Identification (PVI)}
In this task, the LLM agent is given a case with interaction between a customer and an agent and must determine if any company policies have been breached. This involves analyzing the case details and comparing them against policy rules outlined in knowledge articles to identify violations.
\vspace{-5pt}
\paragraph{Knowledge Question Answering (KQA)}
The goal here is for the LLM  agent to answer a specific question based on knowledge articles. This evaluates the agent's capacity to look for accurate and relevant information from the CRM knowledge repository. \looseness=-1

\vspace{-5pt}
\paragraph{Top Issue Identification (TII)}
This task requires the LLM agent to identify the most reported issue for a particular product. %
Given case history, the agent must determine which issue has the highest frequency. This tests the ability to analyze issue reports for trend analysis.
\vspace{-5pt}
\paragraph{Monthly Trend Analysis (MTA)}
The LLM agent must determine which months have the highest number of cases for a given product and a given time period. By analyzing the case history in a given period of time, the LLM agent identifies the month with the most cases%
, demonstrating its ability to recognize trends and patterns over time.
\vspace{-5pt}
\paragraph{Best Region Identification (BRI)}
In this task, the LLM agent's objective is to identify the regions where cases are closed the fastest. The agent must analyze case closure times across various regions and indicate which regions perform best. \looseness=-1

\subsection{Sandbox Environment}
Creating a sandbox environment for \benchmark~ poses unique challenges, particularly related to privacy concerns and the need for realistic data without using real enterprise data.
To this end, we develop a scalable data generation pipeline capable of producing diverse and realistic CRM data across various domains. 
In our setup, we model 16 business objects. %
The complete list of objects and their descriptions can be found in \Cref{apx:sandbox}. %
There are two major challenges for building such a pipeline: object connectivity and hidden casual relationship. In the following subsections, we illustrate how we address these challenges.

\vspace{-5pt}

\paragraph{Object Connectivity} 
Real-world company data is characterized by complex interconnections between objects like \textsc{Case} and \textsc{Account}. Our data generation approach, based on Salesforce's Service Cloud schema, ensures high connectivity. For instance, the \textsc{Case} object is connected to objects like \textsc{Account}, \textsc{Contact}, and \textsc{User}. \Cref{fig:service_erd} displays these interdependencies, creating meaningful data environments. \Cref{tab:dataset_comparison} highlights our benchmark's much higher object connectivity compared to existing work.

\begin{figure}[t!]
    \centering
    \begin{subfigure}[t]{0.48\linewidth}
        \centering
        \includegraphics[width=.95\linewidth, trim=20 20 20 20, clip]{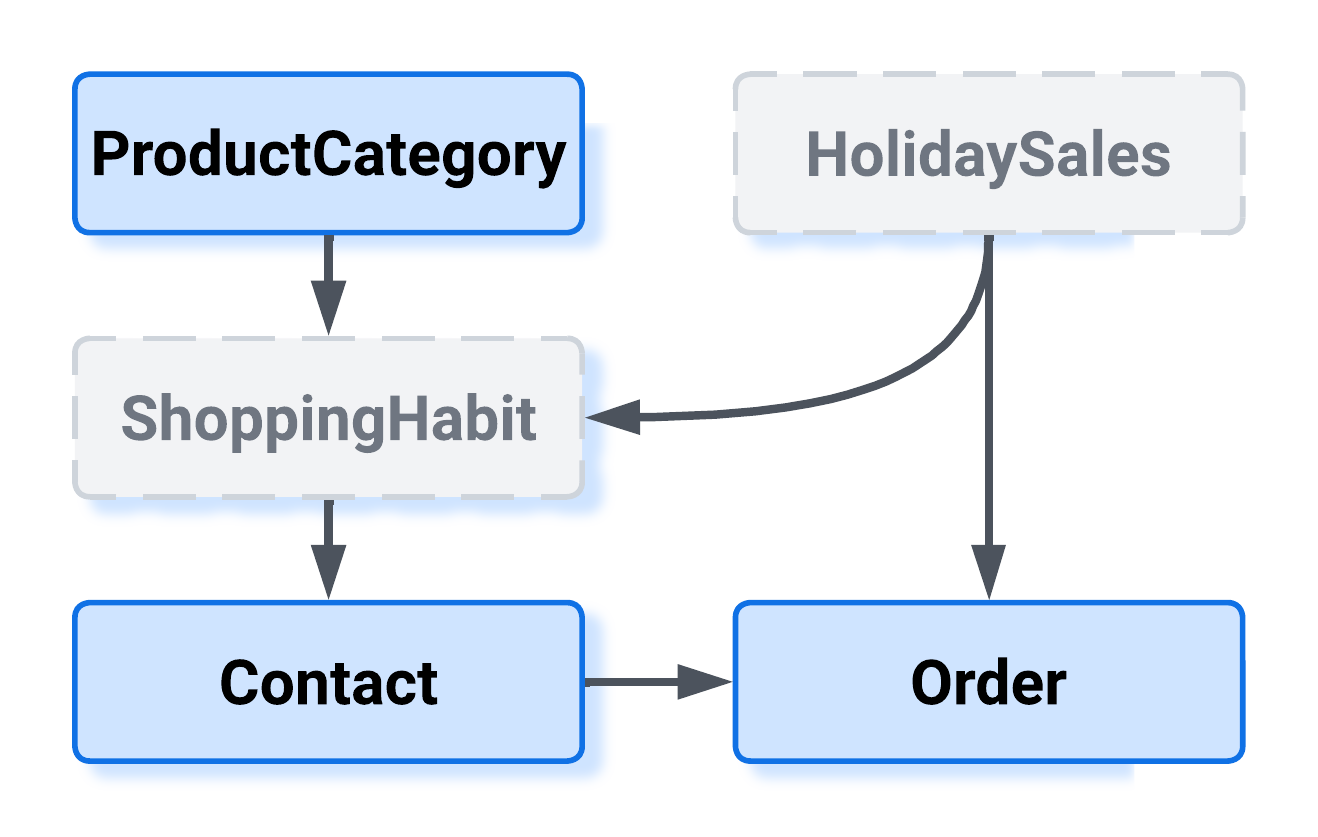}
        \caption{}
    \end{subfigure}%
    ~ 
    \begin{subfigure}[t]{0.48\linewidth}
        \centering
        \includegraphics[width=.95\linewidth, trim=20 20 20 30, clip]{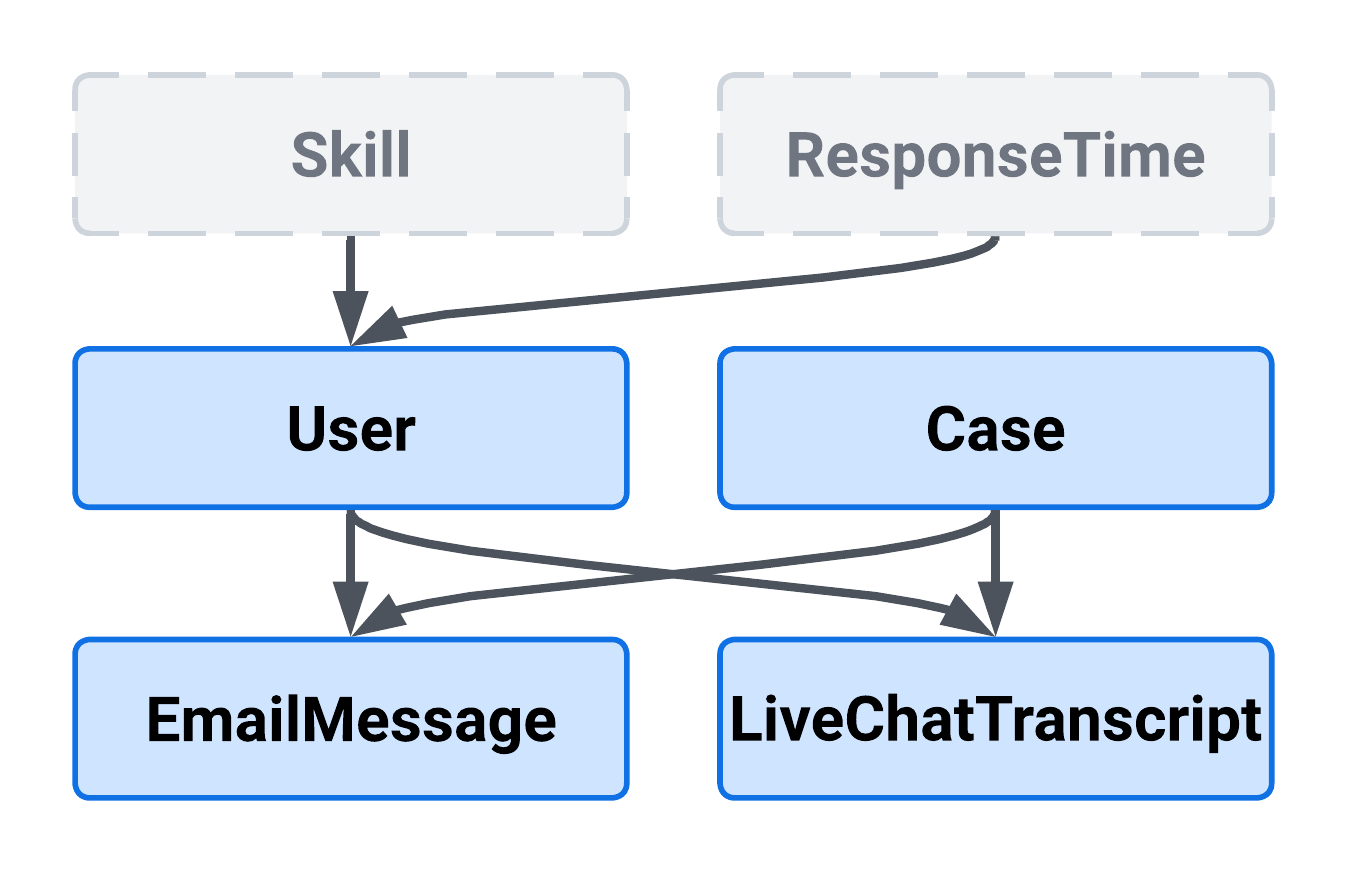}
        \caption{}
    \end{subfigure}
    \vspace{-2mm}
    \caption{Examples of latent variables (gray) influencing object (blue) generation. (a) The \textsc{ShoppingHabit} variable affects when and what type of products a customer buys. (b) The \textsc{Skill} variable determines if an agent can handle a case or needs to transfer it.}
    \label{fig:datagen}
    \vspace{-6mm}
\end{figure}

\vspace{-5pt}
\paragraph{Hidden Causal Relationship}
Replicating the implicit causal relationships found in real-world data presents a significant challenge. To address this, we introduce latent variables that simulate various underlying factors, creating data that mirrors the subtle dependencies and patterns in authentic CRM databases. These latent variables are crucial for facilitating certain tasks and generating desired scenarios. As shown in the example in \Cref{fig:datagen}, the \textsc{ShoppingHabit} variable allows us to more realistically simulate a customer's purchasing patterns based on time periods or holiday seasons. Similarly, the \textsc{Skill} latent variable for the \textsc{User} (Agent) object enables our simulations of \textsc{EmailMessage} and \textsc{LiveChatTranscript} to include situations where an agent is unable to resolve an issue and must transfer it to another agent. Without this latent variable, we would lack scenarios essential for our Transfer Count Understanding task. The full data generation flow is shown in \Cref{fig:data_gen_full}.

\vspace{-5pt}
\paragraph{Quality and Diversity Assurance}

We generate data in JSON format, with each JSON object representing one entry of an object, to ensure higher controllability \cite{huang-etal-2024-embrace, laban-etal-2024-summhay}. Due to the large volume of objects (e.g., 500 \textsc{Product} entries paired with 40+ \textsc{Pricebook} entries resulting in over 20,000 \textsc{PricebookEntry} items) and the limited maximum output tokens of LLMs, directly prompting LLMs to generate all entries of an object is infeasible. To address this, we employ mini-batch prompting with a batch size of 10. However, this approach can lead to duplicated or highly similar content across batches. To mitigate this issue, we implement a two-phase de-duplication strategy. First, for all objects, we include all previously generated entries in the prompt during mini-batch prompting and instruct the LLM not to generate the same content. After data generation, we use string exact matching to remove duplicate entries for fields and objects crucial to certain tasks (e.g., the email of \textsc{User}).

Additionally, we subject the data to a rigorous quality assurance process involving a dual-layer verification. The \textit{format verifier} ensures all data entries conform to predefined schemas by checking whether each entry in the generated mini-batch contains all required fields for the object. Mini-batches that fail this verification are discarded and regenerated. The \textit{content verifier} checks for feasibility for tasks, focusing on objects crucial for specific tasks. For example, in the Named Entity Disambiguation task, we verify that the paraphrased ambiguous product name (1) does not deviate too much from its original name and (2) is not too similar to other products the customer has purchased. In this scenario, the content verifier provides an LLM with a list of products the customer has purchased and the paraphrased product name. If the LLM correctly identifies the product, we retain the entry; otherwise, it is discarded.\footnote{We utilize \texttt{gpt-4o} as the LLM for data synthesis and content verification for its cost efficiency.}

\vspace{-5pt}

\paragraph{Upload to Org} Once the data is generated, we populate it into a clean Simple Demo Org (SDO)\footnote{\url{https://partners.salesforce.com/s/education/general/Salesforce_Orgs}} on Salesforce without latent variables. This exclusion serves two purposes: it mirrors the typical scenario where companies do not have access to such information, thus providing a more realistic testing environment, and it adds an extra layer of challenge compared to testing on the generated databases. Moreover, utilizing Salesforce's SDO as the sandbox eliminates the necessity and complexity for local environment setup, which is required in many related work \cite{styles2024workbench, drouin2024workarena, yao2024tau, zhou2024webarena}. This approach not only facilitates testing but also encourages scientific rigor and future research on our benchmark. The details of the sandbox environment can be found in \Cref{apx:sandbox}.

\vspace{-5pt}

\paragraph{Environment Specification} The input to our data generation pipeline are company name, company description, database schema%
, and the scale of the objects (e.g., number of cases and products). We choose to create an Org for a fictional shoe company due to the diverse product range and customer service scenarios typical in the footwear industry. The scale of our generated data is designed to reflect a mid-sized enterprise, with thousands of orders and hundreds of products and support cases spanning a 4-year period. The total number of entries per object is shown in \Cref{apx:sandbox}.

\vspace{-5pt}
\paragraph{Extensibility}  Our data generation pipeline is designed for flexibility and can be easily adapted to other industries through changes in user-specified input parameters. For instance, by specifying the industry in the company description, our pipeline can automatically generate realistic CRM data tailored to that specific industry, such as finance. %
Furthermore, to accommodate other use cases beyond customer service, such as sales, users would only need to provide the corresponding schema (e.g., Salesforce Sales Cloud schema for sales). This flexibility ensures that the pipeline can be extended to meet a wide range of business needs and LLM agent benchmarking purposes. 

Note that our current setup reflects a simplified version of CRM scenarios, where each \textsc{Case} is linked to both an \textsc{Issue} and a \textsc{Product}. This simplification helps manage the complexity of tasks like Top Issue Identification, which would otherwise require LLM agents to individually analyze every case, making the tasks too infeasible for the current state of LLMs. Our benchmark can be adjusted to create more complex settings by removing such dependencies as LLM capabilities advance.

\subsection{Query Instance Generation}
\label{subsec:query_gen}
Following the creation of the sandbox environment, we generate natural language query instances and their ground-truth answers to benchmark our tasks. For the Knowledge QA tasks, queries can be naively constructed by prompting an LLM each knowledge article to generate question answer pairs~\citep{laban2022discord,huang-etal-2024-embrace}. For the remaining tasks, we construct query instance through a four-step process: (1) seed query construction, (2) ground-truth computation, (3) ID mapping, and (4) query paraphrasing.

We manually create 14 seed queries in total %
with placeholders for corresponding variables, such as time period or product name. This facilitates the development of functions that compute the ground truth answers on the generated database by leveraging the latent variables that are only visible there. %
For example, an agent's policy violation during a live chat is verifiable only within the generated database.
Upon obtaining the answers, we map the IDs in the generated database %
to their counterparts in Salesforce Org, thereby establishing the ground truths for our queries in the sandbox environment. Finally, to ensure diversity in the test queries, we employ an LLM to paraphrase the seed queries, enhancing the robustness and variety of our benchmarking process. An example of this process is shown in the top right of \Cref{fig:project_overview}.

Additionally, to simulate real-world scenarios where some questions may be \textit{unanswerable}, we construct non-answerable queries. Inspired by the non-answerable question schema outlined in \cite{brahman2024art}, we focus on \textit{False Presuppositions} queries, which are most relevant in CRM settings. For example, a query may request the identification of an agent who transfers the most cases during a given period, despite no agents transferring cases in that period. We include non-answerable queries in five tasks: Transfer Time Understanding, Handle Time Understanding, Top Issue Identification, Named Entity Disambiguation, and Policy Violation Identification. For these instances, we expect models to produce ``None'' as outputs. In summary, non-answerable queries account for 30\% of the total queries per corresponding task. Overall, we produce 130 query instances per task, totaling 1,170 queries for \benchmark~. Details and seed queries are provided in \Cref{apx:query_gen}.

\subsection{Tools: APIs and Functions}
Salesforce Orgs naturally support a variety of \textit{general-purpose} APIs, such as the Apex API, REST API, and Tooling API, which are designed to cover a broad set of functionalities within the Salesforce ecosystem. For the scope of our tasks and their integration with a Python environment, we choose to utilize SOQL and SOSL queries\footnote{\url{https://developer.salesforce.com/docs/atlas.en-us.soql_sosl.meta/soql_sosl/sforce_api_calls_soql_sosl_intro.htm}}. SOQL queries are intended for obtaining a specific subset of objects using exact matches or filtering criteria, typically formatted as $\texttt{"SELECT Id ..."}$, while SOSL queries enable fuzzy searching in objects like knowledge articles and product names, formatted as $\texttt{"FIND ..."}$. These two types of queries can theoretically support a wide range of query instances, eliminating the necessity to manually design actions for function calls.

In addition to general-purpose APIs, we also develop \textit{task-specific} tools in the form of Python wrapper functions to facilitate the evaluation of function-calling agents. These functions optimize task performance by providing structured and logical operations directly mapped to typical CRM tasks. We manually define 27 such Python wrapper functions on top of SOQL and SOSL (complete list in \Cref{apx:action_space}) to streamline function calls and target the key components needed for each task. These task-specific functions are designed to maximize reusability across various tasks, minimizing the need for task-specific customizations.

\begin{figure}[t]
    \centering
    \includegraphics[width=0.98\linewidth]{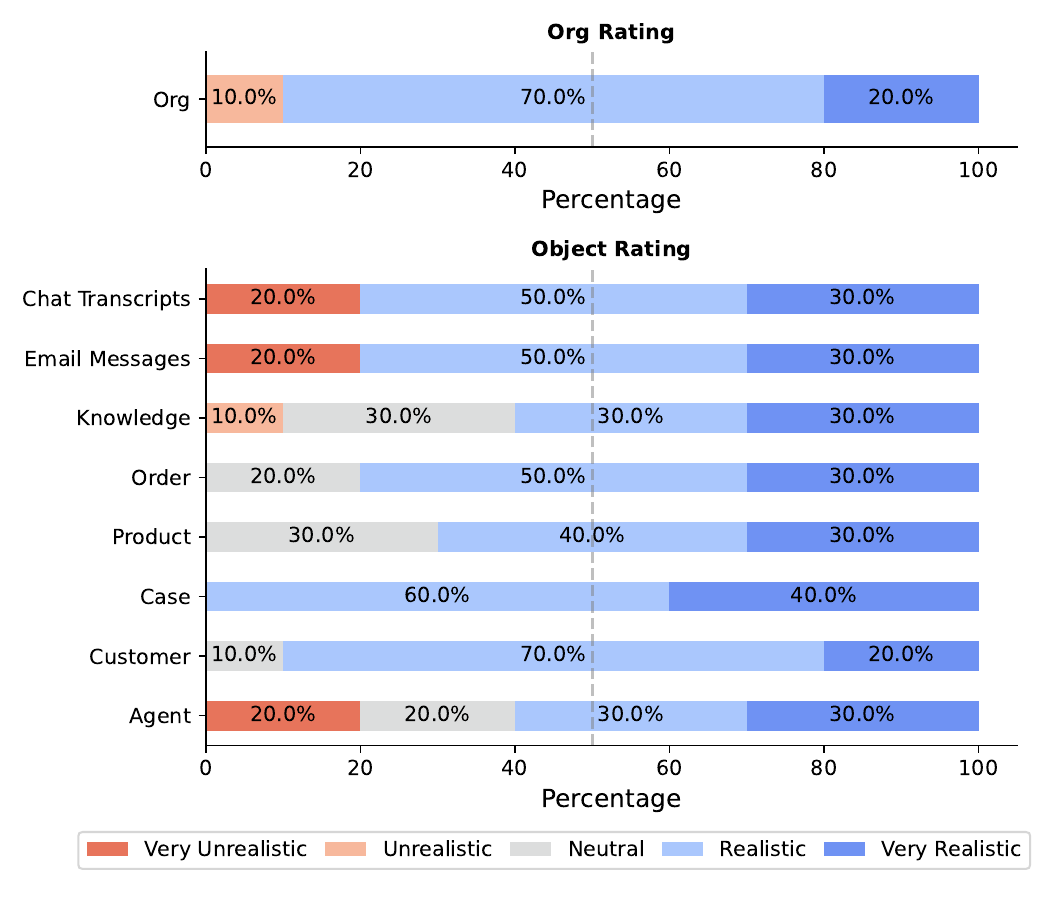}
    \vspace{-3mm}
    \caption{Expert study results. The plots illustrate domain experts' realism ratings for the overall Org structure (top) and individual objects we generated (bottom). \looseness=-1} 
    \vspace{-6mm}    
    \label{fig:expert_study_results}
\end{figure}

\subsection{Expert Study}
\label{subsec:expert_study}
To ensure the realism and practicality of the sandbox environment we developed, we conducted a user study involving ten experts with diverse professional backgrounds who have experience working on Salesforce Orgs daily. These experts were recruited via the User Interviews platform\footnote{\url{https://www.userinterviews.com/}}. Details of the expert study can be found in \Cref{apx:expert_study}.%

Each session of the expert study was structured into three parts. First, we provided the experts with an overview of our sandbox, highlighting key objects such as \textsc{Case} and \textsc{Contact}, and allowing them access through relevant URLs. This initial orientation was designed to familiarize them with the organization. %
Second, we assigned them five query instances sampled from \benchmark~, each representing a different task, to complete. This task completion phase was aimed at evaluating the practical application and operational coherence of the sandbox in executing real-world CRM tasks. Finally, the experts rated the realism of our Org environment compared to the real-world systems they are accustomed to. They also provided detailed rationales for their ratings, giving insights into how our environment aligns with actual CRM scenarios.

The results of our expert study are presented in \Cref{fig:expert_study_results}. The findings are highly encouraging: \textbf{90\% of the experts rated our populated Org as either \textit{Realistic} or \textit{Very Realistic}.} This positive assessment extended to the individual objects within the Org, with more than 77\% of experts giving them similarly high ratings for realism. These results strongly suggest that our sandbox environment closely mirrors real-world CRM systems. We provide the qualitative feedback and rationale from the experts we interviewed in \Cref{tab:expert_study_rationales}.

\section{Benchmarking Experiments}

\subsection{Experimental Settings}
\paragraph{Models} We evaluate state-of-the-art proprietary and open-source LLMs, including \modelname{gpt} models (\modelname{gpt-4o} and \modelname{gpt-3.5-turbo}); \modelname{claude} models (\modelname{claude-3.5-sonnet} and \modelname{claude-3-sonnet}), and the \modelname{llama} models (\modelname{llama-3.1-405b} and \modelname{llama-3.1-70B} \cite{dubey2024llama}).\footnote{We observed the native function-calling mode to perform poorly and hence report the prompt mode performance for Llama 3.1 models.} Additionally, we tested inference-time scaling models for enhanced reasoning capabilities, including \modelname{o1} and \modelname{deepseek-r1} \cite{guo2025deepseek}. 
With these models, we tested three common agentic frameworks: Act, ReAct \cite{yao2023react}, and Function Calling (FC). ReAct is a prompt-based method, with each step characterized by a \textit{thought} and \textit{action} process, while Act is ReAct without the \textit{thought} component. The details of these settings are described in the following paragraphs and \Cref{apx:implementation_details}.%
\looseness=-1%

\paragraph{Action Space}
Every task can be formulated as a Partially Observable Markov Decision Process (POMDP) $(\mathcal{U}, \mathcal{S}, \mathcal{A}, \mathcal{O}, \mathcal{T}, \mathcal{R})$ with instruction space $\mathcal{U}$, state space $\mathcal{S}$, action space $\mathcal{A}$, observation space $\mathcal{O}$, transition function $\mathcal{T}: \mathcal{S} \times \mathcal{A} \to \mathcal{S}$, and reward function $\mathcal{R}: \mathcal{S} \times \mathcal{A} \to \{0, 1\}$. In the Act and ReAct settings, the action space is rich, i.e. $\mathcal{A} = \{\texttt{execute <query>}, \texttt{submit <result>}\}$. Given a user query $u \in \mathcal{U}$ in natural language, an agent can \texttt{execute <query>} to issue a SOQL or SOSL query to interact with the instance to receive the observation $o_t \in \mathcal{O}$ of executing the query in the environment. This continues until the agent chooses to \texttt{submit} and receives a binary reward $r = \mathcal{R}(s_T, \texttt{submit}) \in \{0, 1\}$. In the Function Calling setting, the agent interacts with the environment via API tools implemented as Python functions. In this case the agent is not directly exposed to the Salesforce environment and the object dependencies are kept hidden. Internally the APIs interact with the environment in a controlled manner defined by us. An action $a$ is of the form \texttt{tool\_call\{**kwargs\}}. %
The system prompts for these three setups are described in \Cref{apx:prompts}.

\vspace{-5pt}
\paragraph{Observation Space}
Actions are executed on the sandbox environment through the Simple Salesforce package\footnote{\url{https://github.com/simple-salesforce/simple-salesforce}}. If an action succeeds, the environment will return the queried data in the CRM system; otherwise, an error message, such as incorrect function calling parameters, is returned.%

\vspace{-5pt}
\paragraph{Evaluation Metrics} For the knowledge QA task, since it is an open-ended text generation task, we use F1 scores. %
 For the remaining tasks, we only need to compare the predicted and ground truth object IDs; therefore, an exact match is used.

\begin{table*}[t!]
\vspace{-3mm}
\centering
 \small
 \renewcommand\arraystretch{0.7} %
 
 \resizebox{1.0\linewidth}{!}{
    \begin{tabular}{l|ccccccccc|c}
    \toprule
    \header{Model}  & \header{NCR} & \header{HTU} & \header{TCU} & \header{NED} & \header{PVI} & \header{KQA} & \header{TII} & \header{MTA} & \header{BRI} & \header{ALL}   \\ 
    \midrule
    \multicolumn{11}{l}{\hfill \textit{Act}} \\
    \midrule
    \modelname{gpt-4o} & \gca{43.1} & \gca{10.0} & \gca{17.7} & \gca{30.8} & \gca{28.5} & \textsc{29.3} &  \gca{68.5}  & \gca{29.2} & \gca{7.7} & \gcb{29.4}\\
    \modelname{gpt-4o-mini} & \gca{0.8}	& \gca{38.5} &	\gca{23.8}	& \gca{9.2} & 	\gca{0.0} & \gca{43.1} & \gca{26.9} &	\gca{3.8} & \gca{3.8}	& \gcb{16.7} \\
    \modelname{claude-3.5-sonnet} & \gca{78.5} & \gca{24.6} & \gca{15.4} & \gca{51.5} & \gca{28.5} & \gca{44.7} & \gca{45.4} & \gca{20.8} & \gca{26.9} & \gcb{37.4}\\
    \modelname{claude-3-sonnet} & \gca{9.2} & \gca{26.9} & \gca{24.6} & \gca{30.8} & \gca{23.8} & \gca{16.6} & \gca{16.2} & \gca{1.5} & \gca{0.0} & \gcb{16.6}\\
    \modelname{llama3.1-405b} & \gca{46.2} & \gca{17.7} & \gca{17.7} & \gca{13.9} & \gca{30.0} & \gca{47.0} & \gca{15.4} & \gca{5.4} & \gca{6.9} & \gcb{22.2}\\
    \modelname{llama3.1-70b}  & \gca{28.5} & \gca{20.0}  & \gca{24.6} & \gca{6.2} & \gca{30.0} & \gca{47.9} & \gca{8.5} & \gca{0.0} & \gca{1.5} & \gcb{18.6}\\
    \modelname{llama3.1-8b} & 0.0 & 3.1 & 0.0 & 6.2 & 4.6 & 4.5 & 2.3 & 0.0 & 1.5 & 2.5 \\

    \midrule
    \multicolumn{11}{l}{\hfill \textit{ReAct}} \\
    \midrule

    \modelname{gpt-4o}  & \gca{70.0} & \gca{39.2} & \gca{22.3} & \gca{30.8} & \gca{35.4} & \gca{50.2} & \gca{64.6} & \gca{20.9} & \gca{10.8} & \gcb{38.2} \\
    \modelname{gpt-4o-mini} & \gca{40.8} & \gca{36.9} & \gca{25.4} & \gca{31.5} & \gca{24.6} & \gca{52.8} & \gca{30.0} & \gca{6.2} & \gca{6.2} & \gcb{28.3}\\
    \modelname{claude-3.5-sonnet} & \gca{62.9} & \gca{20.0} & \gca{11.5} & \gca{52.3} & \gca{30.0} & \gca{45.0} & \gca{43.9} & \gca{20.8} & \gca{21.5} & \gcb{34.3}\\
    \modelname{claude-3-sonnet} & \gca{7.7} & \gca{24.6} & \gca{26.9} & \gca{29.2} & \gca{28.5} & \gca{16.0} & \gca{22.3} & \gca{0.8} & \gca{0.0} & \gcb{17.3} \\
    \modelname{llama3.1-405b} & \gca{81.5} & \gca{22.3} & \gca{15.4} & \gca{33.9} & \gca{34.6} & \gca{55.3} & \gca{34.6} & \gca{13.9} & \gca{13.1} & \gcb{33.8}\\
    \modelname{llama3.1-70b} & \gca{48.5} & \gca{20.0} & \gca{13.9} & \gca{33.1} & \gca{37.7} & \gca{48.7} & \gca{23.9} & \gca{13.9} & \gca{10.8} & \gcb{27.8}\\
    \modelname{llama3.1-8b} & 0.0 & 0.0 & 1.5 & \gca{6.2} & \gca{15.4} & \gca{4.0}  & 0.0 & 0.0 & 0.8 & 3.1 \\
    \modelname{o1} & \gca{70.0} & \gca{51.5} & \gca{54.6} & \gca{34.6} & \gca{30.0} & \gca{58.8} & \gca{81.5} & \gca{75.4} & \gca{63.1} & \gcb{57.7}\\
    \modelname{deepseek-r1} & \gca{53.8} & \gca{23.1} & \gca{30.1} & \gca{40.8} & \gca{34.6} & \gca{61.2} & \gca{46.9} & \gca{3.1} & \gca{22.3} & \gcb{35.1}\\
    \midrule
    \multicolumn{11}{l}{\hfill \textit{Function Calling}} \\
    \midrule 
    
    \modelname{gpt-4o} & \gca{60.0} & \gca{47.7} & \gca{81.5} & \gca{46.2} & \gca{39.2} & \gca{30.4} & \gca{97.7} & \gca{27.7} & \gca{59.2} & \gcb{54.4}\\
    \modelname{gpt-4o-mini} & \gca{0.8} & \gca{10.8} & \gca{10.8} & \gca{17.7} & \gca{13.8} & \gca{39.7} & \gca{60.0} & \gca{0.0} & \gca{21.5} & \gcb{19.5} \\
    \modelname{claude-3.5-sonnet} & \gca{4.6} & \gca{33.1} & \gca{82.3} & \gca{52.3} & \gca{30.0} & \gca{40.5} & \gca{69.2} & \gca{26.9} & \gca{36.9} & \gcb{41.8} \\
    \modelname{claude-3-sonnet} & \gca{0.8} &  \gca{1.5} & \gca{30.0} & \gca{25.4} & \gca{41.5} & \gca{23.2} & \gca{12.3} &  \gca{1.5} &  \gca{0.0} & \gcb{15.1}\\
    \modelname{llama3.1-405b (prompt)} & \gca{16.2}  &  \gca{31.5} & \gca{64.6} & \gca{50.0} & \gca{26.9} & \gca{47.6} & \gca{95.4}  & \gca{86.9} & \gca{42.3} & \gcb{51.3}\\
    \modelname{llama3.1-70b (prompt)} & \gca{1.5} & \gca{23.1} & \gca{44.6} & \gca{53.8} &  \gca{37.4} & \gca{42.4} & \gca{93.8} & \gca{43.8} &  \gca{29.2} & \gcb{41.1}\\
    \modelname{llama3.1-8b (prompt)} & 0.0 & 0.0 & 0.0 & 0.0 & 0.0 & 0.0 & 0.0 & 0.0 & 0.0 & 0.0\\
    \modelname{o1 (prompt)} & \gca{60.8} & \gca{68.5} & \gca{66.9} & \gca{60.0} & \gca{24.6} & \gca{39.2} & \gca{99.2} & \gca{84.6} & \gca{74.8} & \gcb{64.3}\\
    \modelname{deepseek-r1 (prompt)} & 0.8 & 0.8 & 2.3 & 0.8 & \gca{24.6} & \gca{34.6} & 0.0 & 13.8 & 3.1 & 9.0\\
    
    \bottomrule
    \end{tabular}
    }
    \vspace{-1mm}
    \caption{Overall performance (\%) of various LLMs under different agentic frameworks on \benchmark~. The evaluation metric is F1 score for the knowledge question answering (KQA) task and exact match for all other tasks. ALL represents the average performance across all tasks.}
\vspace{-3mm}
\label{tab:main_results}
\end{table*}

\subsection{Results}
The main results are summarized in \Cref{tab:main_results}. We made the following observations. First, \textbf{real-world CRM tasks remain challenging for top LLM agents}. Using the ReAct framework, the best model (\texttt{o1}) only achieves an overall score of 57.7\%. Even when equipped with human-crafted functions, the overall performance is still only 64.3\%. These findings highlight the challenges of our \benchmark~. Second, \textbf{stronger and weaker LLMs show opposite trend on different agentic frameworks}. In particular, models like \texttt{gpt-4o} and \texttt{claude-3.5-sonnet} score higher in the FC setting, while their weaker counterparts performs worse when equipped with function calling capabilities. This indicate that human-defined functions may not always help LLM agents, as weaker models may not be able to properly utilize the functions, resulting in lower performance. An intriguing exception is \modelname{deepseek-r1}. Though \modelname{deepseek-r1} is recognized as a strong reasoning model, its tool-calling capabilities seem lacking, primarily due to its (1) inadequate adherence to user instructions and (2) poor ability to adjust previous responses based on external feedback. \textbf{function calling might be unnecessary with a sufficiently strong reasoning model}, as evidenced by \texttt{o1} in the ReAct setting outperforming all other models in the FC setting. Nevertheless, integrating human-crafted functions can still offer performance benefits to strong reasoning models like \texttt{o1}. Finally, \textbf{open-source models are catching up the proprietary LLMs}. Across three settings, we see the \texttt{llama} models score similar, and sometimes higher, than the \texttt{gpt} and \texttt{claude} models. This indicate a closing gap between the open and closed-source models. From \Cref{fig:reward_turn}, we observe how \texttt{llama} models tend to show higher scope for error recovery based on execution feedback than the closed-source models.

\begin{table}[t]
    \small
    \centering
    \begin{adjustbox}{max width=0.49\textwidth}
    {
    \begin{tabular}{llrrr}
        \toprule
        
         & \textbf{Model}  & \textbf{\# Completion Tokens} & \textbf{\# Turns} & \textbf{Cost (\$)} \\
        \midrule
        \multirow{3}{*}{ReAct} & \texttt{gpt-4o} & 48,568.73 & 5.4 & 0.182\\
         & \texttt{claude-3.5-sonnet} & 70,814.75 & 6.9 & 0.228 \\
         & \texttt{llama3.1-405b} & 35,647.29 & 7.3 & 0.125\\
         \midrule
        \multirow{ 2}{*}{FC} & \texttt{gpt-4o} & 78,305.38 & 6.8 & 0.305\\
        & \texttt{claude-3.5-sonnet} & 105,248.43 & 8.1 & 0.371 \\

        \bottomrule
    \end{tabular}
    }
    \end{adjustbox}
    \vspace{-1mm}
    
    \caption{The cost of top-performing agents averaged across queries and tasks. }
    \label{tab:cost_summary}
    \vspace{-4mm}
\end{table}

\subsection{Discussions}

\paragraph{What is the most cost-effective solution?}
Excluding the two reasoning models, in two-third of the agentic frameworks, \texttt{gpt-4o} performs the best. The efficiency of \texttt{gpt-4o} is also reflected in \Cref{tab:cost_summary}, which shows that \texttt{gpt-4o} has the lowest cost per instance and requires the least number of turns to complete a query. Therefore, the most cost-effective solution is using \texttt{gpt-4o} under the function calling setting.

\vspace{-3pt}
\paragraph{How does the type of function affect model performance?} %

In \Cref{tab:main_results}, we observe that equipping LLM agents with function calling capabilities does not necessary results in increased performance. To better understand this phenomenon, we categorizes the functions based on two dimensions: functionality and functional dependency. %
Functionality refers to whether the function solely queries the CRM system via SOSL or SOQL (\textsc{Query}) or if it includes additional operations such as mathematical calculations or aggregations (\textsc{Calculation}). Functional dependency, on the other hand, classifies functions into those that rely on the outputs of other functions (\textsc{Dependent})and those that are independent (\textsc{Independent}). This is crucial because our benchmark requires LLM agents to perform a sequence of calls, with each call dependent on the output of the previous ones \cite{qin2023toolllmfacilitatinglargelanguage,lu2024toolsandboxstatefulconversationalinteractive}. \Cref{tab:ablation_functions_tasks} shows the list of functions and tasks we tested.%

We sampled four function-task pairs from each category to evaluate the performance of \texttt{gpt-4o}, \texttt{gpt-4o-mini}, and  \texttt{claude-3-sonnet} when specific functions were removed from the toolset, substituting two generic functions, \texttt{execute\_soql} and \texttt{execute\_sosl}, to execute arbitrary queries. The findings, summarized in \Cref{tab:ablation_experiments}, indicate that while all function types enhance \texttt{gpt-4o}'s performance, they do not have the same effect on \texttt{gpt-4o-mini} or \texttt{claude-3-sonnet}. This suggests that stronger models are better at utilizing human-crafted functions effectively, whereas weaker models might struggle. Interestingly, \textsc{Calculation} functions, hypothesized to benefit LLMs weak in mathematical operations, may actually decrease performance in weaker models due to their limited function calling capabilities.

\begin{table}[t]
    \small
    \centering
    \begin{adjustbox}{max width=0.49\textwidth}
    {
    \begin{tabular}{llrrr}
        \toprule
        
        \textbf{Functionality} & \textbf{Dependency} & \textbf{\texttt{gpt-4o}} & \textbf{\texttt{gpt-4o-mini}} & \textbf{\texttt{claude-3-sonnet}}\\
        \midrule
        \textsc{Query} & \textsc{Independent} & -6.6 & -6.9 & 2.3\\
        \textsc{Query} & \textsc{Dependent} & -2.9 & 3.0 & 7.5\\
        \textsc{Calculation} & \textsc{Independent} & -9.4 & 4.6 & -3.3\\
        \textsc{Calculation} & \textsc{Dependent} & -26.7 & 4.0 & 3.3\\

        \bottomrule
    \end{tabular}
    }
    \end{adjustbox}
    \vspace{-1mm}
    
    \caption{Performance difference (\%) when removing each category of functions. A lower number indicates more useful functions to the LLM agents. }
    \label{tab:ablation_experiments}
    \vspace{-4mm}
\end{table}

\begin{figure}[t]
    \centering
    \includegraphics[width=0.98\linewidth]{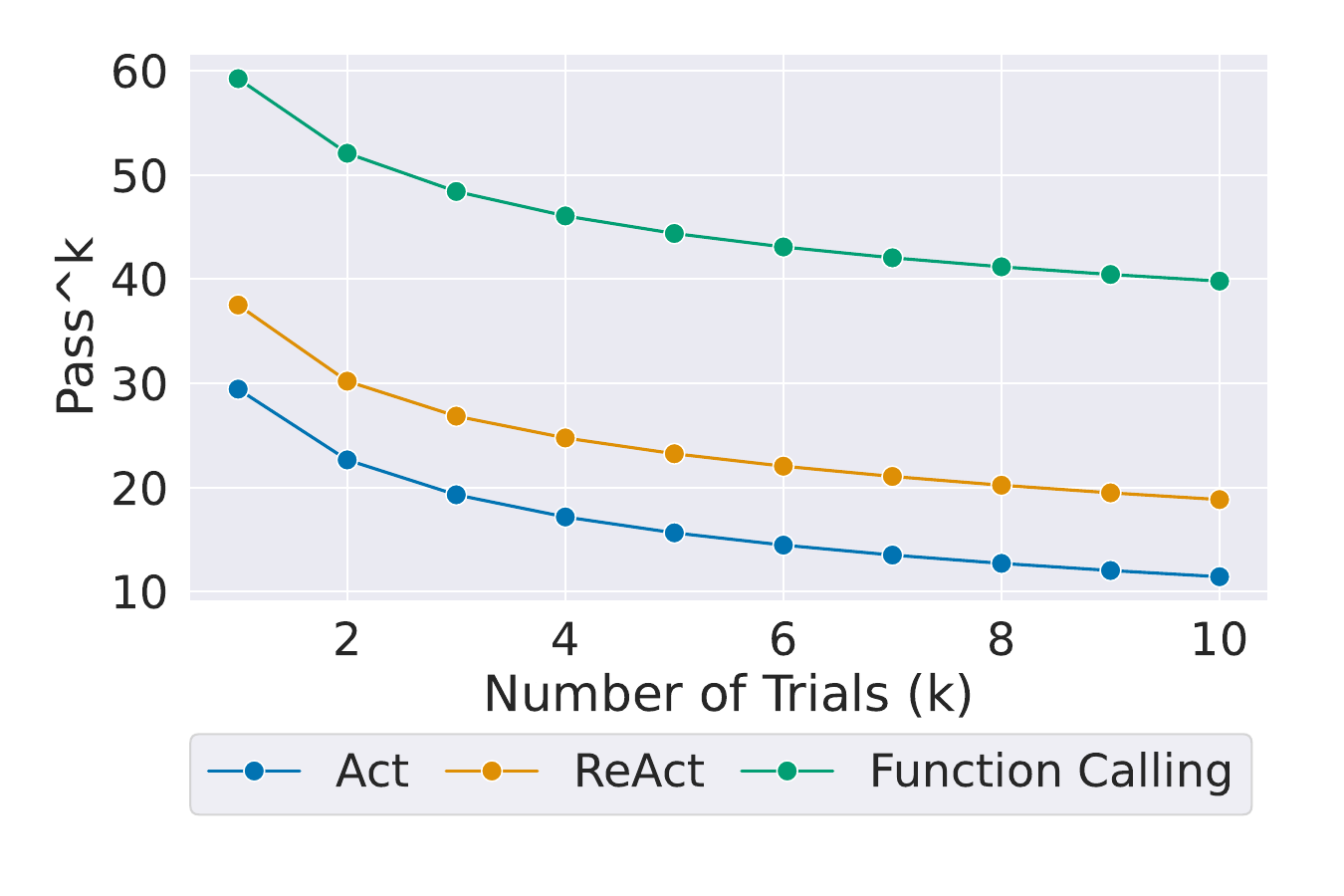}
    \vspace{-3mm}
    \caption{The consistency of \texttt{gpt-4o} using different agentic frameworks.} 
    
    \label{fig:consistency_study}
    \vspace{-5mm}
\end{figure}

\paragraph{How consistent are the agents across multiple trials?}
Consistency is important for LLM agents, especially when deployed in work environments. We evaluate the consistency of LLM agents through multiple trials of prompting. Here, we adapt the \text{pass\textasciicircum k} metric proposed by \citet{yao2024tau}. \text{pass\textasciicircum k} computes the probability that all k independent and identically distributed task attempts are successful, averaged over all tasks. We run ten trials across all tasks in \benchmark~ except for KQA, as the reward for KQA is not binary. The results are shown in \Cref{fig:consistency_study}, we found that, surprisingly, \text{pass\textasciicircum k} for all three agentic frameworks we tested drop at the nearly same rate as k increases. This indicates that the consistency for these three frameworks are similar and that the top-performing LLM cannot reliably solve the tasks with any of the three agentic frameworks we evaluated. \looseness=-1

\section{Related Work}
\paragraph{Agent Benchmark} Several benchmarks have been developed to evaluate LLM-based agents \cite{yao2022webshop, liu2024agentbench, jimenez2024swebench}. Recently, major efforts have focused specifically on web agents, which challenges LLMs to navigate and perform actions on websites. These websites are often about everyday scenarios, such as e-commerce, and social discussion form \cite{deng2023mindweb, he-etal-2024-webvoyager,zhou2024webarena, lu2024weblinx, yoran2024assistantbenchwebagentssolve}. Another line of work focus on evaluating the safety of deploying agents \cite{ruan2024toolemu, yuan2024r, da-etal-2024-safe, qiu2024evaluating}.

\paragraph{Work-oriented Datasets} A few studies have developed datasets specifically for work-oriented tasks. The CRM Benchmark \cite{salesforce2023llmbenchmark} aims to assess LLMs' text generation and summarization abilities in business applications. WorkBench \cite{styles2024workbench} consists of five databases designed to evaluate LLM agents' performance in simple work tasks, such as sending emails, creating calendar invites, and counting traffic sources for a website. $\tau$-Bench \cite{yao2024tau} creates tasks that require interactions with users to obtain relevant information and authorization, achieved by using LLMs to simulate users. WorkArena \cite{drouin2024workarena} builds a web-based work environment that allows for testing agents with visual capabilities.

\section{Conclusion}

This work introduces \benchmark~, a novel benchmark for evaluating LLM agents in performing realistic CRM tasks within professional work environments. By incorporating expert-validated tasks and modeling intricate data interconnections typical of CRM systems, \benchmark~ offers a comprehensive and realistic challenge for LLM agents. Our experiments demonstrate that even state-of-the-art LLMs struggle with these realistic tasks, achieving limited success rates even with function-calling capabilities. These findings highlight the gap between current LLM capabilities and the requirements of real-world CRM scenarios. %
\benchmark~ serves as a foundational step towards more sophisticated evaluations of LLM agents in realistic work environments. \looseness=-1

\section{Ethical Considerations}

This work introduces a benchmark for evaluating LLM agents within the context of CRM systems.  While the data used is synthetically generated, it is modeled after real-world CRM data structures and tasks.  Thus, it is important to consider the ethical implications of this work, particularly regarding data biases and privacy concerns.

\paragraph{Data Bias} Although synthetic, the data is generated by models trained on real-world data, which may contain inherent biases. These biases, related to customer demographics, purchase behavior, or case resolution, could be inadvertently reflected in the generated data, potentially perpetuating stereotypes or discriminatory practices. Thankfully, after conducting a thorough manual inspection of the generated data to identify potential biases, we did not observe such patterns.

\paragraph{Privacy Concerns} While our benchmark does not use any real customer data and therefore does not have access to personal information, the structure and nature of CRM data itself can raise privacy concerns.  The tasks in our benchmark involve accessing sensitive customer information, albeit synthetic. To ensure responsible handling of this data, even though synthetic, we performed a thorough manual inspection to verify the absence of any personally identifiable information and to confirm that the data cannot be used to infer private information about individuals. This meticulous review process reinforces our commitment to ethical data practices and mitigates potential privacy risks.

\section{Limitations}

The \benchmark~ comprises nine tasks that thoroughly assess the ability of LLM agents to perform duties typically associated with three primary roles within a realistic environment: Service Manager, Service Agent, and Service Analyst. Nonetheless, this study does not encompass other common personas in CRM, such as sales representatives. We aim to incorporate these additional roles in our future studies.

\bibliography{custom}

\appendix

\clearpage
\appendix

\section{Further Discussions}

\paragraph{Reward vs number of turns}

\begin{figure}[t]
    \centering
    \includegraphics[width=0.98\linewidth]{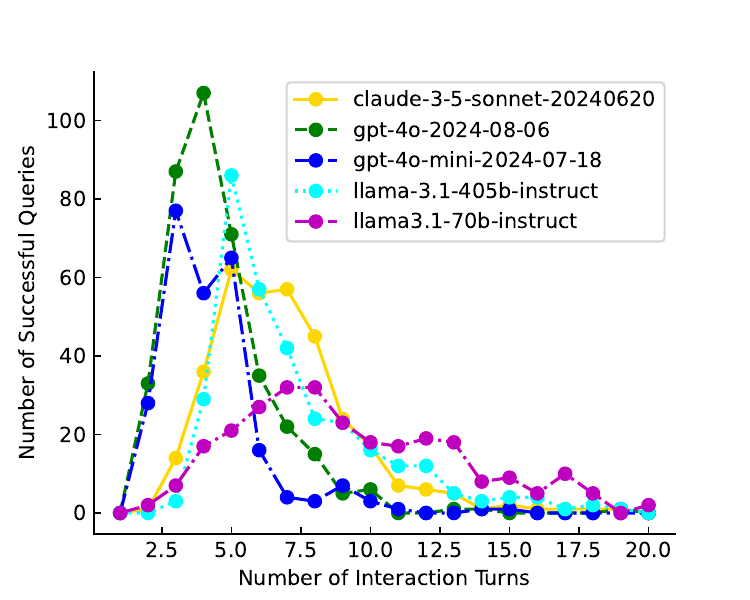}
    
    \caption{Distribution of the number of turns it takes for agents to reach the goal under \textit{ReAct}.} 
    
    \label{fig:reward_turn}
\end{figure}

In \Cref{fig:reward_turn}, we show the distribution of the number of turns it takes for agents to successfully complete an user query.

\paragraph{Non-answerable query analysis}
In \Cref{tab:unanswerable_queries}, we present the performance of each LLM agents. Overall, LLM agents are good at handling such queries, compared to standard queries. Interestingly, a trend shown in \Cref{tab:main_results} is observed in this experiment as well: function calling only benefit stronger LLMs, while weaker LLMs like \modelname{claude-3-sonnet} and \modelname{gpt-4o} performs worse when equipped with function calling capabilities.

\section{Query Generation Details}
\label{apx:query_gen}
\Cref{tab:seedqueries} show the complete list of seed queries used in our experiments. More examples of how the final queries are constructed can be found in \Cref{tab:query_examples}.

\section{Action Space Details}
\label{apx:action_space}
In the text-based agent settings (i.e. ReAct and Act), the actions include (1) executing SOSL queries, (2) executing SOQL queries, and (3) submitting the answer. In the function-calling settings, the actions are a list of carefully designed functions, shown in \Cref{tab:functions}.

\begin{table*}[t]
\vspace{-3mm}
\centering
 \small
 
 \resizebox{0.65\linewidth}{!}{
    \begin{tabular}{l|ccccc}
    \toprule
    \header{Model}  & \header{HTU} & \header{TCU} & \header{NED} & \header{TII}  & \header{PVI}      \\ 
    \midrule
    \multicolumn{6}{l}{\hfill \textit{Act}} \\
    \midrule
    \modelname{gpt-4o} &\gca{ 15.4} & \gca{48.7}  & \gca{ 94.9} & \gca{ 87.2} & \gca{ 92.3}\\
    
    \modelname{gpt-4o-mini} & \gca{ 94.9} & \gca{ 79.5} & \gca{ 30.8} & \gca{ 79.5} & \gca{ 74.4}\\

    \modelname{claude-3.5-sonnet} & \gca{25.6} & \gca{28.2} & \gca{82.1} & \gca{33.3} & \gca{84.6} \\
    \modelname{claude-3-sonnet} & \gca{ 84.6} & \gca{ 79.5} & \gca{ 100.0} & \gca{ 51.3} & \gca{ 74.4}\\
    
    \modelname{llama3.1-405b} & \gca{ 56.4} & \gca{ 51.3} & \gca{ 46.2} & \gca{ 38.5} & \gca{0.0}\\
    \modelname{llama3.1-70b} & \gca{ 46.2} & \gca{ 76.9} & \gca{ 20.5} & \gca{ 20.5} & \gca{ 100.0}\\

    \midrule
    \multicolumn{6}{l}{\hfill \textit{ReAct}} \\
    \midrule
    
   \modelname{gpt-4o} & \gca{ 64.1} & \gca{ 48.7} & \gca{ 100.0} & \gca{ 84.6} & \gca{ 74.4}\\
    
    \modelname{gpt-4o-mini} &\gca{ 97.4} & \gca{ 82.1} & \gca{ 97.4} & \gca{ 61.5} & \gca{ 71.8}\\

    \modelname{claude-3.5-sonnet} & \gca{ 12.8} & \gca{ 7.7} & \gca{ 87.2} & \gca{ 30.8} & \gca{ 82.1} \\
    \modelname{claude-3-sonnet} & \gca{ 79.5} & \gca{ 84.6} & \gca{ 94.9} & \gca{ 69.2} & \gca{ 94.9}\\
    
    \modelname{llama3.1-405b} & \gca{ 53.8} & \gca{ 38.5} & \gca{ 97.4} & \gca{ 41.0} & \gca{ 64.1} \\
    \modelname{llama3.1-70b} & \gca{ 64.1} & \gca{ 41.0} & \gca{ 97.4} & \gca{ 17.9} & \gca{ 17.9}\\

    \midrule
    \multicolumn{6}{l}{\hfill \textit{Function Calling}} \\
    \midrule 
    \modelname{gpt-4o} & \gca{ 59.0} & \gca{ 84.6} & \gca{ 74.4} & \gca{ 100.0} & \gca{ 35.9}\\
    
    \modelname{gpt-4o-mini} & \gca{ 15.4} & \gca{ 7.7} & \gca{ 0.0} & \gca{ 0.0} & \gca{ 0.0}\\
    \modelname{claude-3.5-sonnet} & \gca{ 52.6} & \gca{ 74.4} & \gca{ 100.0} & \gca{ 100.0} & \gca{ 100.0}\\
    \modelname{claude-3-sonnet} & \gca{ 2.6} & \gca{ 15.4} & \gca{ 59.0} & \gca{ 38.5} & \gca{ 56.4}\\
    
    \bottomrule
    \end{tabular}
    }
    \vspace{-2mm}
    \caption{Performance (\%) of various LLMs under different agentic frameworks on \benchmark~ for the non-answerable queries.}
\vspace{-6mm}
\label{tab:unanswerable_queries}
\end{table*}

\begin{table*}[t!]
\vspace{-3mm}
\centering
 \small
 \renewcommand\arraystretch{0.9} %
 
 \resizebox{1.0\linewidth}{!}{
    \begin{tabular}{l}
        \toprule
        \texttt{1. In [YEAR] [MONTH/QUARTER/SEASON], identify the agent who managed more than [NCASES] cases and had the [EXTREMA] handle time.} \\
        \texttt{2. In the past [TIMEPERIOD], find the agent with the [EXTREMA] handle time among those who managed more than [NCASES] cases.} \\
        \texttt{3. During the last [TIMEPERIOD], which agent had the [EXTREMA] average handle time for those handling over [NCASES] cases?} \\
        \texttt{4. In [YEAR] [MONTH/QUARTER/SEASON], identify the agent who managed more than [NCASES] cases and had the [EXTREMA] transfer counts.} \\
        \texttt{5. In the past [TIMEPERIOD], find the agent with the [EXTREMA] transfer counts among those who managed more than [NCASES] cases.}\\
        \texttt{6. During the last [TIMEPERIOD], which agent had the [EXTREMA] average transfer counts for those handling over [NCASES] cases?}  \\
        \texttt{7. Which knowledge article did the agent violate policy?}\\
        \texttt{8. Today is [TODAY]. Is there any month in which the cases we received for [PRODUCT] is much more than other months over the past [TIMEPERIOD]?} \\
        \texttt{9. Today is [TODAY]. For [PRODUCT], what is the most common issue in the last [TIMEPERIOD].}\\
        \texttt{10. Today is [TODAY]. In [YEAR] [MONTH/QUARTER/SEASON], what is the most common issue for [PRODUCT].}\\
        \texttt{11. Today is [TODAY]. In which states do we close cases the fastest in the last [TIMEPERIOD]?} \\
        \texttt{12. Today is [TODAY]. In [YEAR] [MONTH/QUARTER/SEASON], which states do we close cases the fastest.} \\
        \texttt{13. What is the best agent to assign to for this case?} \\
        \texttt{14. Today is [TODAY]. Show me the [PRODUCT] that I ordered [PERIOD] ago.} \\
        \bottomrule
    \end{tabular}
    }
    \vspace{-2mm}
    \caption{The full set of seed queries used for query generation.}
\vspace{-3mm}
\label{tab:seedqueries}
\end{table*}

\begin{table*}[t]
    \small
    \centering
    \begin{adjustbox}{max width=0.98\textwidth}
    {
    \begin{tabular}{ll}
        \toprule
        
        \textbf{Functions} & \textbf{Description}\\
        \midrule
        \texttt{get\_agents\_with\_max\_cases(subset\_cases)} & Returns a list of agent IDs with the maximum number of cases from the given subset of cases. \\
        
        \texttt{get\_agents\_with\_min\_cases(subset\_cases)} & Returns a list of agent IDs with the minimum number of cases from the given subset of cases. \\
        
        \texttt{calculate\_average\_handle\_time(cases)} & Calculates the average handle time for each agent based on a list of cases. \\
        
        \texttt{get\_start\_date(end\_date, period, interval\_count)} & Calculates the start date based on the end date, period, and interval count. \\
        
        \texttt{get\_period(period\_name, year)} & Calculates the start and end date based on the period name and year. \\
        
        \texttt{get\_agent\_handled\_cases\_by\_period(start\_date, end\_date)} & Retrieves the number of cases handled by each agent within a specified time period. \\
        
        \texttt{get\_qualified\_agent\_ids\_by\_case\_count(agent\_handled\_cases, n\_cases)} & Filters agent IDs based on the number of cases they have handled. \\
        
        \texttt{get\_cases(start\_date, end\_date, agent\_ids, case\_ids,} & Retrieves cases based on various filtering criteria. \\
        \texttt{\hspace{9pt}order\_item\_ids, issue\_ids, statuses)} & \\
        
        \texttt{get\_non\_transferred\_case\_ids(start\_date, end\_date)} & Retrieves the IDs of cases that were not transferred between agents within a specified date range. \\
        
        \texttt{get\_agent\_transferred\_cases\_by\_period(start\_date, end\_date,} & Retrieves the number of cases transferred between agents within a specified date range. \\
        \texttt{\hspace{9pt}qualified\_agent\_ids)} & \\
        
        \texttt{get\_shipping\_state(cases)} & Adds shipping state information to the given cases. \\
        
        \texttt{calculate\_region\_average\_closure\_times(cases)} & Calculates the average closure times for cases grouped by region (shipping state). \\
        
        \texttt{get\_order\_item\_ids\_by\_product(product\_id)} & Retrieves the order item IDs associated with a given product. \\
        
        \texttt{get\_issue\_counts(start\_date, end\_date, order\_item\_ids)} & Retrieves the issue counts for a product within a given time period. \\
        
        \texttt{find\_id\_with\_max\_value(values\_by\_id)} & Identifies the ID with the maximum value from a dictionary. \\
        
        \texttt{find\_id\_with\_min\_value(values\_by\_id)} & Identifies the ID with the minimum value from a dictionary. \\
        
        \texttt{get\_account\_id\_by\_contact\_id(contact\_id)} & Retrieves the Account ID associated with a given Contact ID. \\
        
        \texttt{get\_purchase\_history(account\_id, purchase\_date, related\_product\_ids)} & Retrieves the purchase history for a specific account, date, and set of products. \\
        
        \texttt{get\_month\_to\_case\_count(cases)} & Counts the number of cases for each month from a list of cases. \\
        
        \texttt{search\_knowledge\_articles(search\_term)} & Searches for knowledge articles based on a given search term. \\
        
        \texttt{search\_products(search\_term)} & Searches for products based on a given search term. \\
        
        \texttt{get\_issues()} & Retrieves a list of issue records. \\

        \texttt{get\_email\_messages\_by\_case\_id(case\_id)} & Retrieves the email exchanges for a given case. \\

        \texttt{get\_livechat\_transcript\_by\_case\_id(case\_id)} & Retrieves the live chat transcript for a given case. \\
        
        \texttt{submit(content)} & Returns the response content. \\
        
        \bottomrule
    \end{tabular}
    }
    \end{adjustbox}
    
    \vspace{-2mm}
    \caption{The complete list of functions for the function calling settings.} 
    \label{tab:functions}
    \vspace{-5mm}
\end{table*}

\section{Sandbox Environment Details}
\label{apx:sandbox}
\begin{figure*}[t]
    \centering
    \includegraphics[width=0.98\linewidth]{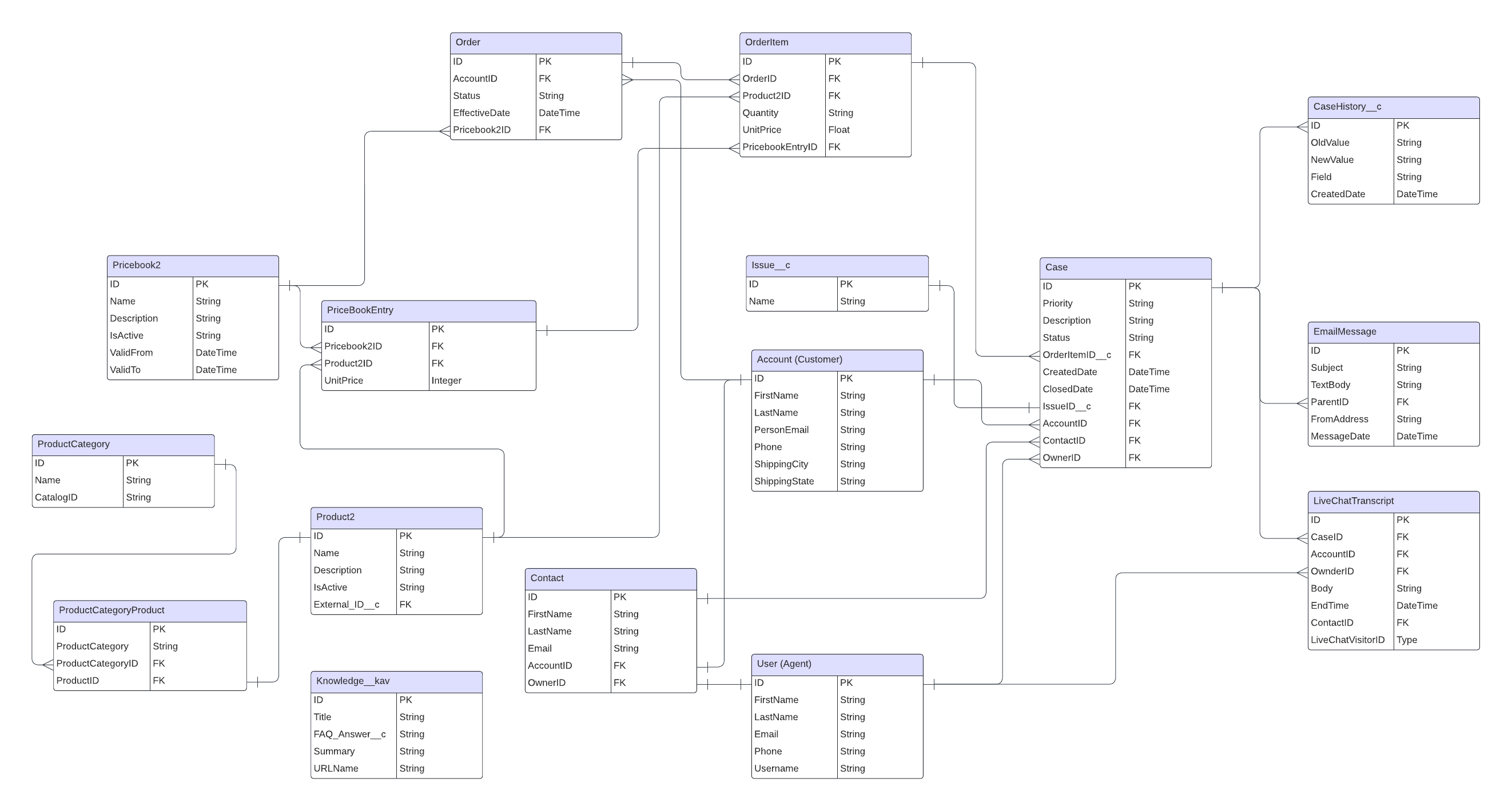}
    
    \caption{The objects and their dependencies in our sandbox environment.} 
    
    \label{fig:service_erd}
\end{figure*}

\begin{figure*}[t]
    \centering
    \includegraphics[width=0.98\linewidth]{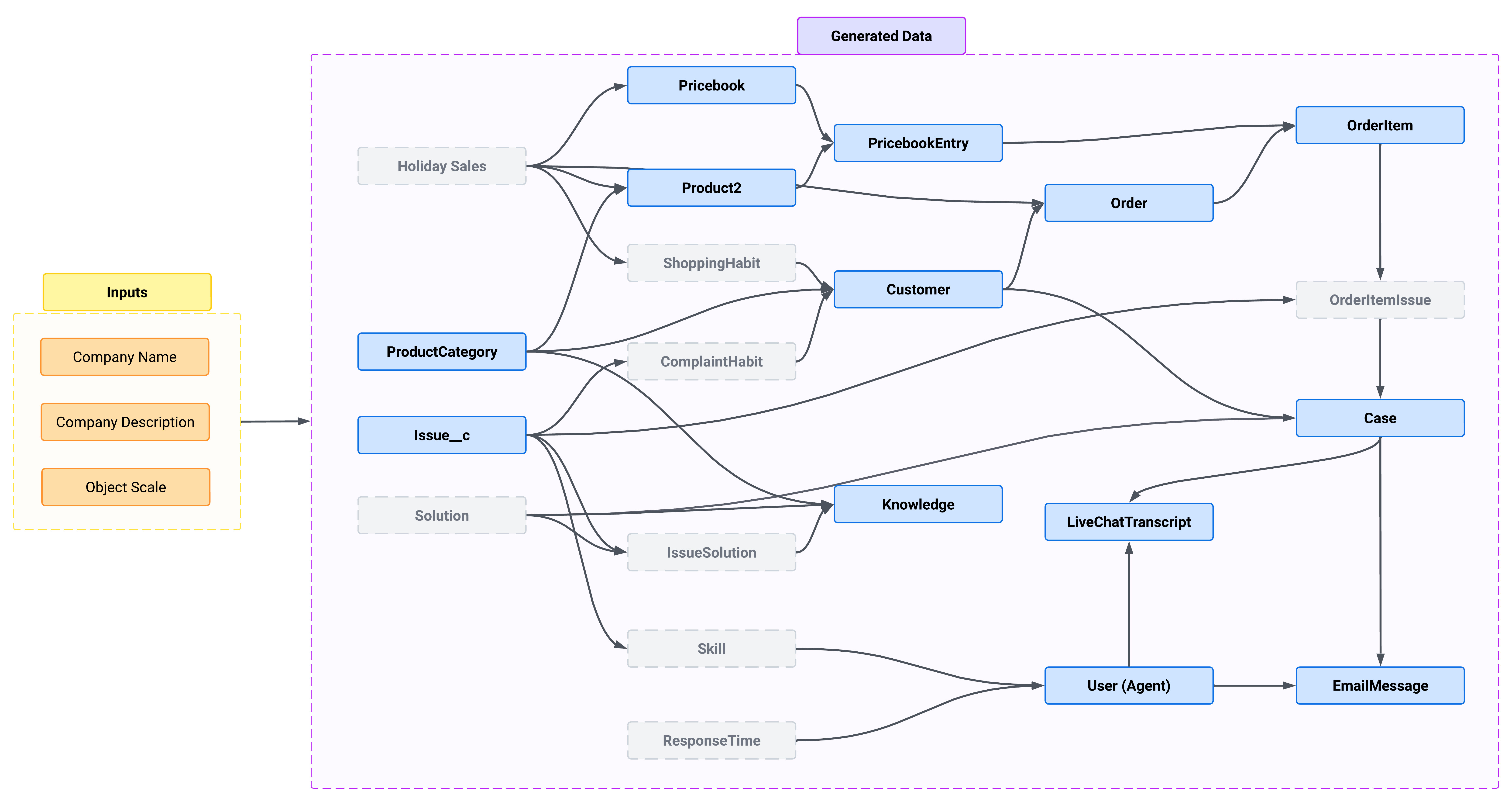}
    
    \caption{Data generation overview. The generation of each object is conditioned on the previously generated objects with arrows pointing to it. Blue boxes represent standard object, while gray boxes denote latent variables that are not uploaded to the Salesforce Org.} 
    
    \label{fig:data_gen_full}
\end{figure*}

We show the objects and dependencies in \Cref{fig:service_erd}. These objects, except for $\texttt{Knowledge\_\_kav}$ are densely connected, reflecting the complexity of real-world work environment. The total number of entry per objects is shown in \Cref{tab:entry_per_object}. Our data generation flow is shown in \Cref{fig:data_gen_full}.

\begin{table}[t]
    \small
    \centering
    \begin{adjustbox}{max width=0.49\textwidth}
    {
    \begin{tabular}{lr}
        \toprule
        
        \textbf{Object} & \textbf{Number of Entries} \\
        \midrule
        \textsc{User} & 100 \\
        \textsc{Contact} & 196 \\
        \textsc{ProductCategory} & 12 \\
        \textsc{Product} & 500 \\
        \textsc{OrderItem} & 71,00 \\
        \textsc{Pricebook} & 44 \\
        \textsc{PricebookEntry} & 22,000\\
        \textsc{Case} & 977 \\
        \textsc{Order} & 2,071 \\
        \textsc{EmailMessage} & 3,234\\
        \textsc{LiveChatTranscript} & 387\\
        \textsc{Knowledge} & 45\\

        \bottomrule
    \end{tabular}
    }
    \end{adjustbox}
    \vspace{-2mm}
    
    \caption{The number of entries per object. }
    \label{tab:entry_per_object}
    \vspace{-5mm}
\end{table}

\begin{table*}[h!]
    \centering
    \begin{minipage}{0.95\textwidth}   
    \centering
    \begin{tcolorbox} 
        \centering
        \small
        
        \begin{tabular}{p{0.95\textwidth}}
        \textbf{Handle Time Understand} \newline
    
        \textbf{Seed query}: In [YEAR] [MONTH/QUARTER/SEASON], identify the agent who managed more than [NCASES] cases and had the [EXTREMA] handle time. \newline

        \textbf{Filled-in query}: In 2021 February, identify the agent who managed more than 2 cases and had the highest handle time. \newline

        \textbf{Paraphrased query}: In February 2021, determine the agent with the longest handle time who managed more than 2 cases. \\
        
        \midrule
        \textbf{Top Issue Identification} \newline
        
        \textbf{Seed query}: In [YEAR] [MONTH/QUARTER/SEASON], what is the most common issue for [PRODUCT]? \newline

        \textbf{Filled-in Query}: In 2023 Q2, what is the most common issue for Flex Yoga Mat? \newline
        
        \textbf{Paraphrased query}: What was the most frequent issue with Flex Yoga Mat in the second quarter of 2021?\\
        \end{tabular}
        \end{tcolorbox}
        \vspace{-4mm}
        \caption{Examples of the query generation process.}
        \vspace{-3mm}
        \label{tab:query_examples}
    \end{minipage}
    \end{table*}

\subsection{Object Details}

Below, we describe the details of each object.

\begin{itemize}
    \item \textbf{ProductCategory}: Represents the category that products are organized in.
    \item \textbf{Product2}: Represents a product that your company sells.
    \item \textbf{ProductCategoryProduct}: Holds the relation between product and product category to assign products to a category. 
    \item \textbf{Pricebook2}: Represents a price book that contains the list of products.
    \item \textbf{Pricebook Entry}: Represents a product entry (an association between a Pricebook2 and Product2) in a price book.
    \item \textbf{Order}: Represents an order associated with a contract or an account.
    \item \textbf{Order Item}: Represents an order product that the company sells.
    \item \textbf{Knowledge}: Documentation or information articles that are accessible to users or customers.
    \item \textbf{Contact}: Refers to an individual or party related to an account.
    \item \textbf{Issue}: Represents a type of problem raised by a customer.
    \item \textbf{Account}: An entity, company, or individual your company does business with. In B2C setting, an account represents a customer.
    \item \textbf{User (agent)}: System user, often representing customer support agents.
    \item \textbf{Case}: A record that describes a customer inquiry or issue.
    \item \textbf{CaseHistory}: A log of the changes and updates made to a case over time.
    \item \textbf{EmailMessage}: Email communication related to cases or customer inquiries between an agent and a customer.
    \item \textbf{LiveChatTranscript}: A conversation from a live chat session between an agent and a customer.
\end{itemize}

\section{Prompts}
\label{apx:prompts}

In this section, we display the prompts used in our experiments. \Cref{tab:act_prompts}, \Cref{tab:react_prompts}, \Cref{tab:fc_prompts}  show the system prompt for the Act, ReAct, and Function Calling settings, respectively.

\begin{table*}[h!]
    \centering
    \begin{minipage}{0.95\textwidth}   
    \centering
    \begin{tcolorbox} 
        \centering
        \small
        
        \begin{tabular}{p{0.95\textwidth}}
        
    $\text{You are an expert in Salesforce and you have access to a Salesforce instance.}$\\ 
       $\textbf{ Instructions}$
    
    $\text{- You will be provided a question, the system description, and relevant task context.}$
    
    $\text{- Interact with the Salesforce instance to build Salesforce Object Query Language}$
    
    $\text{(SOQL) or Salesforce Object Search Language (SOSL) queries as appropriate, to help you answer the question.}$
    
    $\text{- Salesforce Object Search Language (SOSL) can be used to construct text-based search queries}$
    
    $\text{against the search index.}$
    
    $\text{- Your generation should always be an Action command and NOTHING ELSE.}$
    
    $\text{Generate only one Action command.}$
    
    $\text{- DO NOT generate ANY system observation, you will receive this based on your Action command.}$
    
    $\text{- If no record is found matching the requirements mentioned, just return 'None'.}$
    
    $\text{Here is a description of how to use these commands:}$
    
    $\textbf{ Action}$
    
    $\text{- Can be 'execute' or 'submit'.}$
    
    $\text{- \underline{execute}, to execute SOQL/SOSL that will return the}$ 
    
    $\text{observation from running the query on the Salesforce instance.}$
    
    $\text{- \underline{submit}, to return the final answer of the task to the user.}$
    
    $\text{- Format: <execute> a valid SOQL/SOSL query </execute> or <submit> response to user </submit>}$
    
    $\textbf{ Guidelines}$
    
    $\text{- Execute SOQL/SOSL queries to understand the Salesforce instance}$
    
    $\text{that will help you find the answer to the question.}$
    
    $\text{- When you are confident about the answer, submit it.}$
    
    $\text{- Always end with a submit action containing ONLY the answer, NO full sentences or any explanation.}$
    
    $\textbf{ Example 1}$
    
    $\text{Question: What is the total number of opportunities?}$
    
    $\text{Output:}$

    $\text{<execute> SELECT COUNT() FROM Opportunity </execute>}$
    
    $\text{     (If the observation from the Salesforce instance 100, your next step can be)}$
    
    $\text{<submit> 100 </submit> NOT <submit> The total number of opportunities is 100 </submit>}$
    
    $\textbf{ Example 2}$\newline
    
    \textit{[... Hide details for space... ]}\newline
    
    $\textbf{ Salesforce instance description}$
    
    $\text{The objects available in the Salesforce instance are:}$
    
    $\text{User, Account, Contact, Case, Knowledge\_\_kav, ProductCategory, Product2, ...}$
    
    $\textbf{ The fields available for the objects along with their descriptions and dependencies are:}$
    
    $\text{User}$
    
    $\text{   - FirstName: First name of the agent}$
    
    $\text{   - LastName: Last name of the agent}$
    
    $\text{   - Email: Email address of the agent}$
    
    \textit{[... Hide details for space... ]}\newline

    $\textbf{ Additional task context}$
    
    $\textbf{ Handle/Transfer Times Policies}$
    
    \textit{[... Hide details for space... ]}

        \end{tabular}
        \end{tcolorbox}
        \vspace{-4mm}
        \caption{The system prompt used in the Act setting.}
        \vspace{-3mm}
        \label{tab:act_prompts}
    \end{minipage}
    \end{table*}

\begin{table*}[h!]
    \centering
    \begin{minipage}{0.95\textwidth}   
    \centering
    \begin{tcolorbox} 
        \centering
        \small
        
        \begin{tabular}{p{0.95\textwidth}}
        
        $\text{You are an expert in Salesforce and you have access to a Salesforce instance.}$\\ 
       $\textbf{ Instructions}$
    
    $\text{- You will be provided a question, the system description, and relevant task context.}$
    
    $\text{- Think step by step and interact with the Salesforce instance to build Salesforce Object Query Language}$
    
    $\text{(SOQL) or Salesforce Object Search Language (SOSL) queries as appropriate, to help you answer the question.}$
    
    $\text{- Salesforce Object Search Language (SOSL) can be used to construct text-based search queries}$
    
    $\text{against the search index.}$
    
    $\text{- Your generation should always be a Thought followed by an Action command and NOTHING ELSE.}$
    
    $\text{Generate only one Thought and one Action command.}$
    
    $\text{- DO NOT generate ANY system observation, you will receive this based on your Action command.}$
    
    $\text{- If no record is found matching the requirements mentioned, just return 'None'.}$
    
    $\text{Here is a description of how to use these commands:}$
    
    $\textbf{Thought}$
    
    $\text{- A single line of reasoning to process the context and inform the decision making. }$
    
    $\text{Do not include any extra lines.}$
    
    $\text{- Format: <thought> your thought </thought>}$
    
    $\textbf{ Action}$
    
    $\text{- Can be 'execute' or 'submit'.}$
    
    $\text{- \underline{execute}, to execute SOQL/SOSL that will return the}$ 
    
    $\text{observation from running the query on the Salesforce instance.}$
    
    $\text{- \underline{submit}, to return the final answer of the task to the user.}$
    
    $\text{- Format: <execute> a valid SOQL/SOSL query </execute> or <submit> response to user </submit>}$
    
    $\textbf{ Guidelines}$
    
    $\text{- Always start with a Thought and then proceed with an Action.}$
    
    $\text{- Generate only one Thought and one Action command at a time.}$
    
    $\text{- Execute SOQL/SOSL queries to understand the Salesforce instance}$
    
    $\text{that will help you find the answer to the question.}$
    
    $\text{- When you are confident about the answer, submit it.}$
    
    $\text{- Always end with a submit action containing ONLY the answer, NO full sentences or any explanation.}$
    
    $\textbf{ Example 1}$
    
    $\text{Question: What is the total number of opportunities?}$
    
    $\text{Output:}$
    
    $\text{<thought> I need to find the total number of opportunities in the system. </thought>}$
    
    $\text{<execute> SELECT COUNT() FROM Opportunity </execute>}$
    
    $\text{     (If the observation from the Salesforce instance 100, your next step can be)}$
    
    $\text{<thought> I have found the total number of opportunities. </thought>}$
    
    $\text{<submit> 100 </submit> NOT <submit> The total number of opportunities is 100 </submit>}$
    
    $\textbf{ Example 2}$\newline
    
    \textit{[... Hide details for space... ]}\newline
    
    $\textbf{ Salesforce instance description}$
    
    $\text{The objects available in the Salesforce instance are:}$
    
    $\text{User, Account, Contact, Case, Knowledge\_\_kav, ProductCategory, Product2, ...}$
    
    $\textbf{ The fields available for the objects along with their descriptions and dependencies are:}$
    
    $\text{User}$
    
    $\text{   - FirstName: First name of the agent}$
    
    $\text{   - LastName: Last name of the agent}$
    
    $\text{   - Email: Email address of the agent}$
    
    \textit{[... Hide details for space... ]}\newline

    $\textbf{ Additional task context}$
    
    $\textbf{ Handle/Transfer Times Policies}$
    
    \textit{[... Hide details for space... ]}

        \end{tabular}
        \end{tcolorbox}
        \vspace{-4mm}
        \caption{The system prompt used in the ReAct setting.}
        \vspace{-3mm}
        \label{tab:react_prompts}
    \end{minipage}
    \end{table*}

\begin{table*}[h!]
    \centering
    \begin{minipage}{0.95\textwidth}
    \centering
    \begin{tcolorbox}
        \centering
        \small

        \begin{tabular}{p{0.95\textwidth}}

    $\textbf{Instructions}$

    $\text{- You are an expert in Salesforce and you have access to a Salesforce instance.}$

    $\text{- You will be provided a question, the system description, and relevant task context.}$

    $\text{- Interact with the Salesforce instance using the tools provided to help you answer the question.}$

    $\text{- You should ALWAYS make ONLY ONE tool call at a time.}$ 
    
    $\text{If you want to submit your final answer, use the 'submit' tool.}$

    $\text{If not, you should call some other tool. But ALWAYS make a tool call.}$

    $\text{- Always end by calling 'submit' tool containing ONLY the answer, NO full sentence or any explanation.}$

    $\text{- If your answer is empty that is there are no records found matching the requirements mentioned,}$
    
    $\text{just return 'None' to the 'submit' tool.}$

    $\textbf{Additional task context}$

    $\textbf{Case Routing Policy}$

    $\text{The case routing policy determines the best agent to assign the given new case based on the following criteria}$

    $\text{- Issue Expertise: The agent who has closed the most cases}$
    
    $\text{associated with the issue most similar to the given case.}$

    $\text{- Product Expertise: If there is a tie in issue expertise, the best agent is the one who has solved the most cases}$

    $\text{associated with the product most relevant to the given case.}$

    $\text{- Workload: If there is still a tie, the best agent is the one that has least cases with Status not 'Closed'.}$

    $\textbf{Domain Details}$

    $\textbf{Quarters of the Year}$

    $\text{- Q1: January 1 to March 31 (both inclusive).}$

    $\text{- Q2: April 1 to June 30 (both inclusive).}$

    $\text{- Q3: July 1 to September 30 (both inclusive).}$

    $\text{- Q4: October 1 to December 31 (both inclusive).}$

    $\textbf{Seasons}$

    $\text{- Winter: December 1 to February 28/29 (both inclusive).}$

    $\text{- Spring: March 1 to May 31 (both inclusive).}$

    $\text{- Summer: June 1 to August 31 (both inclusive).}$

    $\text{- Autumn/Fall: September 1 to November 30 (both inclusive).}$

    $\textbf{Time Periods}$

    $\text{- Past 2 quarters: This refers to any timeframe spanning two quarters back from a specified `end\_date`.}$

    $\text{That translates to a six-month period retrospectively from the `end\_date`.}$

    $\text{- Issue Significantly More Than Other Months: This means there is a month where the number of cases}$

    $\text{reported are larger than all other months.}$

        \end{tabular}
    \end{tcolorbox}
    \vspace{-4mm}
    \caption{The system prompt used in the Function Calling setting.}
    \vspace{-3mm}
    \label{tab:fc_prompts}
    \end{minipage}
\end{table*}

\section{Expert Study Details}
\label{apx:expert_study}
\begin{table}[t]
    \small
    \centering
    \begin{adjustbox}{max width=0.49\textwidth}
    {
    \begin{tabular}{llc}
        \toprule
        
        \textbf{Profession} & \textbf{Gender} & \textbf{Age} \\
        \midrule
        Customer Service Associate & Female & 23 \\
        Customer Service Associate & Female & 25 \\
        Customer Service Agent & Male & 39 \\
        Customer Service Associate & Male & 29 \\
        Customer Service Advisor & Male & 49 \\
        Customer Service Manager & Male & 39 \\
        Account Executive & Female& 25 \\
        Technical Support & Female& 38 \\
        Customer Service Advisor & Female & 25 \\
        Customer Service Agent & Female & 35 \\

        \bottomrule
    \end{tabular}
    }
    \end{adjustbox}
    \vspace{-2mm}
    
    \caption{The background of the participants in our expert study. }
    \label{tab:expert_study_participants}
    \vspace{-5mm}
\end{table}

As detailed in \Cref{tab:expert_study_participants}, we recruited a diverse range of domain experts for our study. The participants varied in age, gender, and professional backgrounds.

\subsection{Recruitment Criteria}

Using the User Interviews platform, we set the job filter such that the participants of our survey must have a job title of one of the following: 
\begin{itemize}
    \item Account Manager
    \item Technical Support Engineer
    \item Support Engineer
    \item Technical Support Specialist
    \item Technical Support Manager
    \item Technical Support Technician
    \item Technical Support Agent
    \item Technical Support Expert
    \item Account Manager/Agent
    \item Account Manager/Analyst
    \item Customer Service Advisor/Customer Service Associate
    \item Customer Service Associate 
    \item Customer Service Representative
\end{itemize}

In addition, we have created a screener survey. The most important question in the survey is ``How often do you use Salesforce CRM?''. The valid candidate must select the option ``Several times a day'' to be able to participate in our study.

\subsection{The study}
We use Google Form to conduct expert studies due to its ease to use. The study is broken down into three parts:

\begin{itemize}
    \item \textbf{Part 1}: Familiarizing the Org [5 minutes]. This is for a broad overview of some of the objects in this Org.
    \item  \textbf{Part 2}: Task Completion [45 minutes]. At this stage, they are be given tasks regarding customer service. They should try to accomplish as many as possible within the 45-minute time frame.
    \item \textbf{Part 3}: Quality Rating [10 minutes]. Based on their experience with the first two parts of this study, rate the quality of the Org and objects.
\end{itemize}

Below, we illustrate how each part is executed.

\paragraph{Part 1}

\begin{figure*}[t]
    \centering
    \includegraphics[width=0.6\linewidth, trim=0 50 0 170, clip]{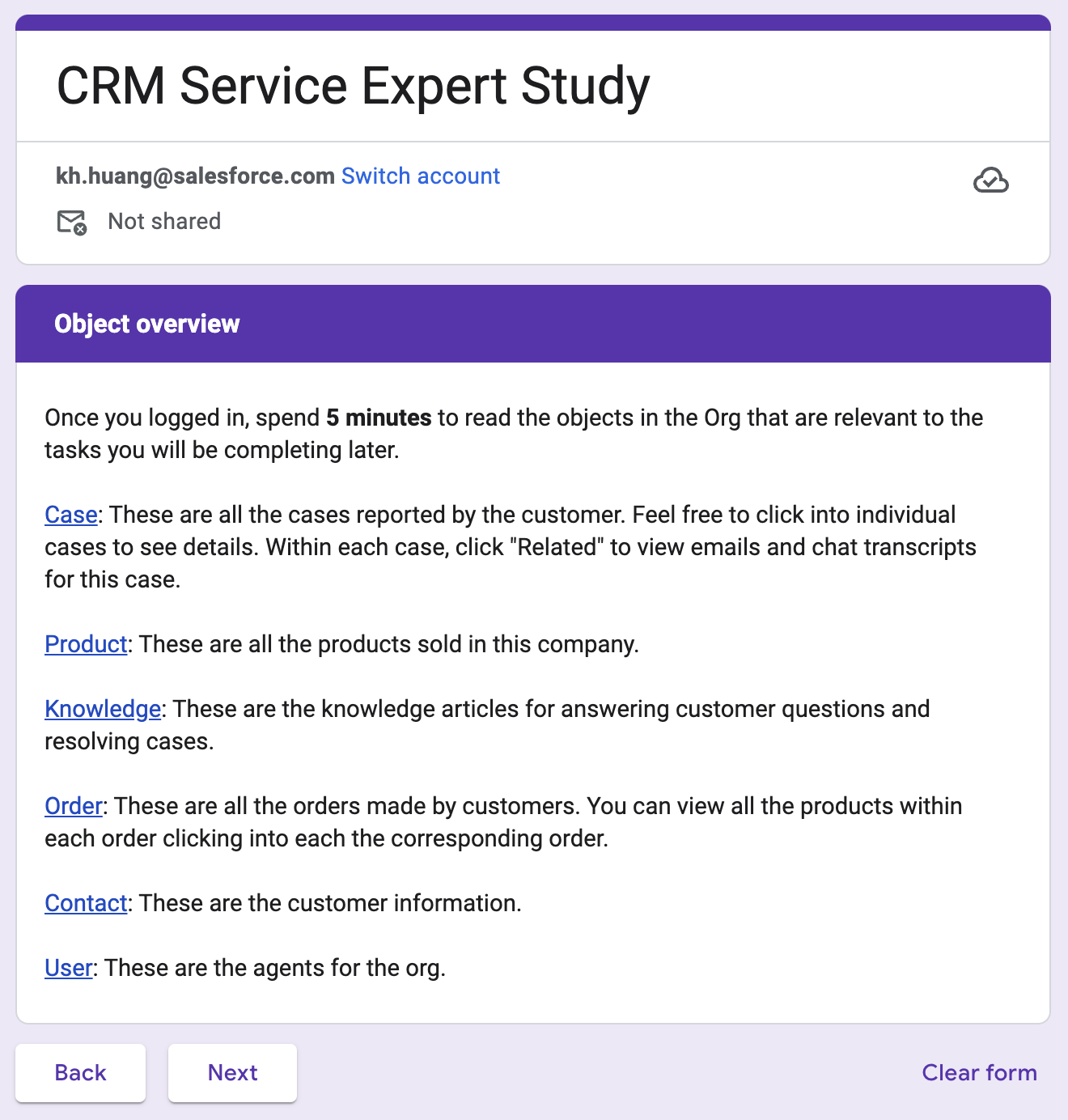}
    
    \caption{The instructions and interface of Part 1 of our expert study.} 
    
    \label{fig:expert_study_part1}
\end{figure*}

In this part, we provide interviewee the log in credentials to our created Org (sandbox environment). Once they log in, they are instructed to spend 5 minutes to read the objects in the Org that are relevant to the tasks they will be completing later. The instructions and interface for this part are shown in \Cref{fig:expert_study_part1}.

\begin{figure*}[b]
    \centering
    \includegraphics[width=0.6\linewidth, trim=0 50 0 190, clip]{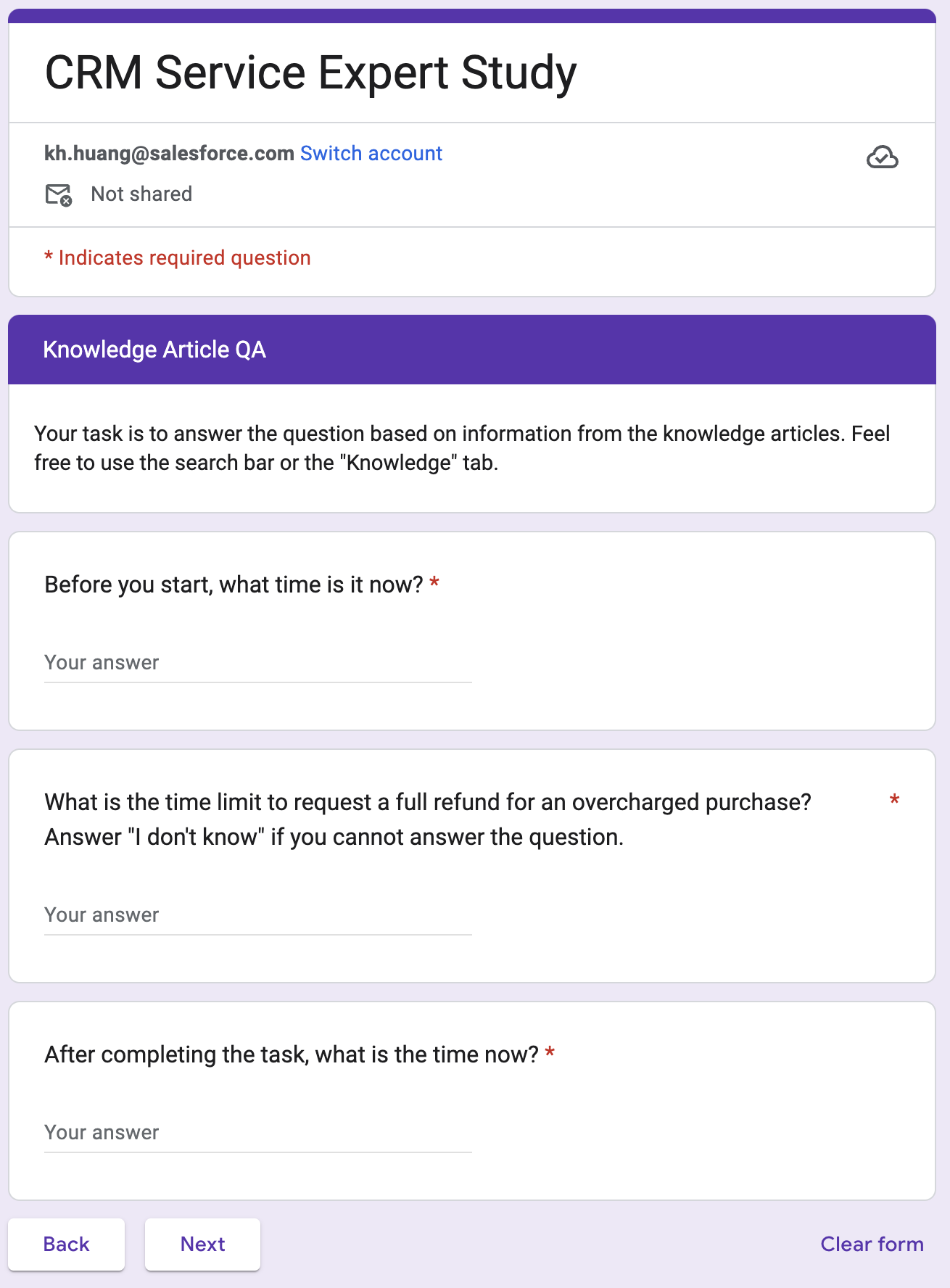}
    
    \caption{An example query instance for the part 2 of expert study.} 
    
    \label{fig:expert_study_part2}
\end{figure*}

\paragraph{Part 2}
After familiarizing with our created Org, participants are then asked to complete the tasks. They are required to complete 5 query instances from \benchmark~. An example of the query is shown in \Cref{fig:expert_study_part2}.

\begin{figure*}[t]
    \centering
    \includegraphics[width=0.6\linewidth, trim=0 50 0 190, clip]{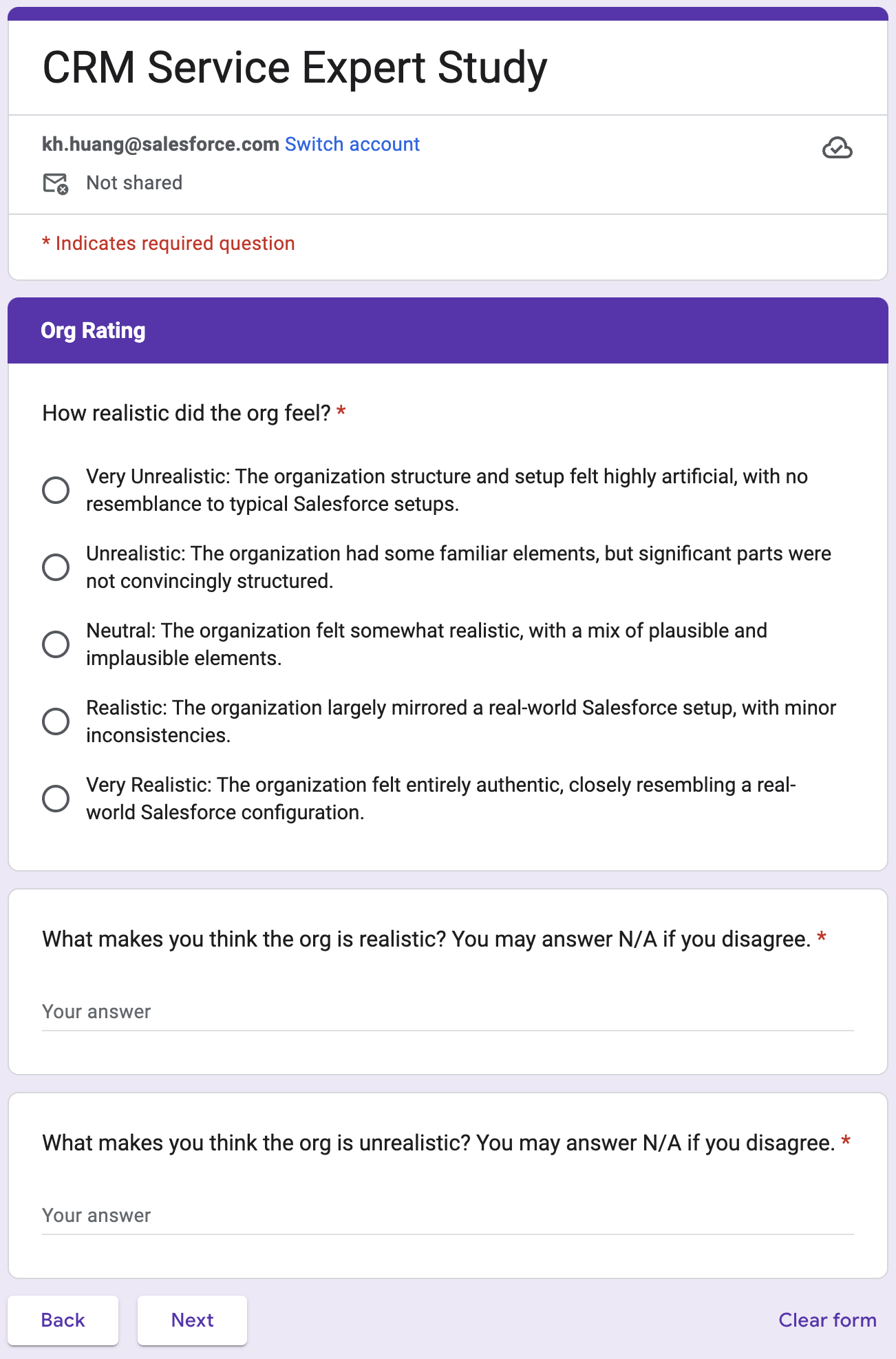}
    
    \caption{An example question for the part 3 of our expert study.} 
    
    \label{fig:expert_study_part3}
\end{figure*}

\paragraph{Part 3}
Upon completing the first two parts of the expert study, in the final stage, participants are asked to rate the realism of our Orgs and data. In addition to providing ratings, they also need to provide rationales for their ratings. An example question is shown in \Cref{fig:expert_study_part3}.

Below, we provide the rating and descriptions for participants to choose from. 

\textbf{Org ratings}:
\begin{itemize}
    \item Very Unrealistic: The organization structure and setup felt highly artificial, with no resemblance to typical Salesforce setups.
    \item Unrealistic: The organization had some familiar elements, but significant parts were not convincingly structured.
    \item Neutral: The organization felt somewhat realistic, with a mix of plausible and implausible elements.
    \item Realistic: The organization largely mirrored a real-world Salesforce setup, with minor inconsistencies.
    \item Very Realistic: The organization felt entirely authentic, closely resembling a real-world Salesforce configuration.
\end{itemize}

\textbf{Object ratings}:
\begin{itemize}
    \item I don’t know/I’m not familiar with the object.
    \item Very Unrealistic: The objects seemed fundamentally flawed or invented with little regard for typical Salesforce objects.
    \item Unrealistic: The objects had recognizable features but were generally not representative of actual Salesforce objects.
    \item Neutral: The objects were moderately realistic, combining elements of both realistic and unrealistic features.
    \item Realistic: The objects were mostly realistic and aligned well with typical objects used in Salesforce, with minor issues.
    \item Very Realistic: The objects felt entirely authentic and perfectly matched real-world Salesforce objects.
\end{itemize}

\subsection{Qualitative Feedback}
In \Cref{tab:expert_study_rationales}, we present qualitative feedback and rationale from the experts we interviewed, as they determine whether our Organization and Object are perceived as Realistic or Unrealistic.

\begin{table*}[t]
    \small
    \centering
    {
    \begin{tabular}{p{0.15\linewidth} p{0.15\linewidth} p{0.6\linewidth}}
        \toprule
        
        \textbf{Rated Instance} & \textbf{Rating} & \textbf{Rationale} \\
        \midrule
        \multirow{5}{*}{Org}  & \multirow{4}{*}{Realistic} & 1. This is really similar to what a normal Salesforce instance looks like (i.e. the one we use at our company). However, there are a few missing details in some of the pages like when you click into a contact or account.  \\
               &  & 2. It feels like my usual Salesforce Dashboard for my current job, I could more or less get a feel for the general navigation of the simulation. \\ 
               & & 3. This is what salesforce looks like for me to find case numbers and information about each of the cases that were indentified by customers.\\
               & & 4. Knowing nothing about the org I was able to fumble my way around and find what I needed to.\\
        \cmidrule{2-3}
         & Unrealistic & 1. The lack of customer data/information filling out the rest of the fields. There is no semblance of a system that's been ``worked in'' and everything feels very empty and confusing with nothing to fill the interface. \\
        \midrule
        \multirow{5}{*}{Object} & \multirow{4}{*}{Realistic} & 1. Case management, customer interactions, knowledge base, and the transcripts were what made it realistic.\\
        & & 2. I think the email correspondence wasn't perfect, but it did feel rather authentic.\\
        & & 3. I feel like the cases and customers issue are real life issue so I feel like they are realistic. \\
        &  & 4. They have similar details and structures as a typical salesforce deployment (at least in my company). A lot of those elements have the same fields that are in their expected places (like Details, additional context on the right side) \\
        \cmidrule{2-3}
        & Unrealistic & 1. The unrealistic ones are finding the agent information. This is unrealistic because I should be able to filter and find each of the agent transfers and handle time with the customers.\\

        \bottomrule
    \end{tabular}
    }
    
    \caption{Example rationales provided by domain experts for their ratings of our sandbox environment's realism.}
    \label{tab:expert_study_rationales}
\end{table*}

\section{Implementation Details}
\label{apx:implementation_details}
We use the OpenAI API for the \modelname{gpt} models; Amazon Bedrock API for the \modelname{claude} models; and the Together API for the \modelname{llama3.1} models. Below we provide the version of the model we tested:
\begin{itemize}
    \item \texttt{o1}: o1-2024-12-17
    \item \texttt{gpt-4o}: gpt-4o-2024-08-06
    \item \texttt{gpt-3.5-turbo}: gpt-3.5-turbo-0125
    \item \texttt{deepseek-r1}: deepseek-ai/DeepSeek-R1
    \item \texttt{claude-3.5-sonnet}: anthropic.claude-3-5-sonnet-20240620-v1:0
    \item \texttt{claude-3-sonnet}: anthropic.claude-3-sonnet-20240229-v1:0
    \item \texttt{llama3.1-405b}: meta-llama/Meta-Llama-3.1-405B-Instruct-Turbo
    \item \texttt{llama3.1-70b}: meta-llama/Meta-Llama-3.1-70B-Instruct-Turbo
    \item \texttt{llama3.1-8b}: meta-llama/Meta-Llama-3.1-8B-Instruct-Turbo
    
\end{itemize}

We choose the ReAct setting over Plan based approaches that decompose the task into more manageable steps as prior works showed that in SQL based database querying tasks, planning strategy is less flexible to altering its plan by adjusting to execution feedback \cite{yang2024intercode}. We set the max actions for each instance to 20, temperature to 0, and top\_p to 1 for all experiments. \looseness=-1

\begin{table*}[t]
    \small
    \centering
    \begin{adjustbox}{max width=0.98\textwidth}
    {
    \begin{tabular}{llcc}
        \toprule
        
        \textbf{Functionality} & \textbf{Dependency} & \textbf{Function} & \textbf{Task} \\
        \midrule
        \multirow{4}{*}{\textsc{Query}} & \multirow{4}{*}{\textsc{Independent}} & \texttt{get\_order\_item\_ids\_by\_product(product\_id)} & MTA \\
        & &\texttt{get\_order\_item\_ids\_by\_product(product\_id)} & NCR\\
        & &\texttt{search\_products(search\_term)}  & NED\\
        & & \texttt{get\_account\_id\_by\_contact\_id(contact\_id)} & NED\\
        \midrule
        \multirow{4}{*}{\textsc{Query}} & \multirow{4}{*}{\textsc{Dependent}} &  \texttt{get\_non\_transferred\_case\_ids(start\_date, end\_date)}& HTU  \\
        & & \texttt{get\_cases(start\_date, end\_date, agent\_ids, case\_ids,} & NCR\\
        & & \texttt{get\_cases(start\_date, end\_date, agent\_ids, case\_ids,} & BRI\\
        & & \texttt{get\_cases(start\_date, end\_date, agent\_ids, case\_ids,} & HTU\\
        \midrule
        \multirow{4}{*}{\textsc{Calculation}} & \multirow{4}{*}{\textsc{Independent}} & \texttt{get\_start\_date(end\_date, period, interval\_count)} & TCU \\
        & & \texttt{get\_start\_date(end\_date, period, interval\_count)} & BRI\\
        & & \texttt{get\_start\_date(end\_date, period, interval\_count)} & TII\\
        & & \texttt{get\_period(period\_name, year)} & TCU \\
        \midrule
        \multirow{4}{*}{\textsc{Calculation}} & \multirow{4}{*}{\textsc{Dependent}} & \texttt{calculate\_region\_average\_closure\_times(cases)}  & BRI \\

        & &  \texttt{get\_qualified\_agent\_ids\_by\_case\_count(agent\_handled\_cases, n\_cases)}  & TCU \\
        & & \texttt{calculate\_average\_handle\_time(cases)}  & HTU \\
        & & \texttt{get\_agents\_with\_max\_cases(subset\_cases)}  & NCR \\

        \bottomrule
    \end{tabular}
    }
    \end{adjustbox}
    \vspace{-2mm}
    
    \caption{The list of functions and tasks tested in \Cref{tab:ablation_experiments}.}
    \label{tab:ablation_functions_tasks}
    \vspace{-5mm}
\end{table*}

\end{document}